\documentclass{clv2}

\usepackage{tabularx}
\usepackage{soul}
\usepackage{amssymb}
\usepackage{amsmath}
\usepackage{url}
\usepackage{color}
\usepackage{subfigure} 
\usepackage{array,multirow,graphicx}
\usepackage{soul}
\usepackage[normalem]{ulem}

\showhyphens{\ul{underlining}}
\showhyphens{\st{overstriking}}

\renewenvironment{quote}
  {\begin{trivlist} \setlength\leftskip{0.5cm} \setlength\rightskip{0.5cm}
   \item\relax}
  {\end{trivlist}}

\newcommand{\argmax}{\mathop{\mathrm{argmax}}\limits}   

\usepackage{cleveref}

\usepackage{xcolor}

\issue{?}{?}{2016}

\runningtitle{Parsing Argumentation Structures}

\runningauthor{Stab and Gurevych}

\begin{document}

\title{Parsing Argumentation Structures in Persuasive Essays}

\author{Christian Stab\thanks{Ubiquitous Knowledge Processing Lab (UKP-TUDA), Department of Computer Science, Technische Universit\"{a}t Darmstadt}}
\affil{Technische Universit\"at Darmstadt}

\author{Iryna Gurevych\thanks{Ubiquitous Knowledge Processing Lab (UKP-TUDA), Department of Computer Science, Technische Universit\"{a}t Darmstadt and Ubiquitous Knowledge Processing Lab (UKP-DIPF), German Institute for Educational Research}}
\affil{Technische Universit\"at Darmstadt and German Institute for Educational Research}

\maketitle

\begin{abstract}
In this article, we present a novel approach for parsing argumentation structures.
We identify argument components using sequence labeling at the token level and apply a new joint model for detecting argumentation structures. 
The proposed model globally optimizes argument component types and argumentative relations using integer linear programming.
We show that our model considerably improves the performance of base classifiers and significantly outperforms challenging heuristic baselines. 
Moreover, we introduce a novel corpus of persuasive essays annotated with argumentation structures.
We show that our annotation scheme and annotation guidelines successfully guide human annotators to substantial agreement. 
This corpus and the annotation guidelines are freely available for ensuring  reproducibility and to encourage future research in computational argumentation.\footnote{www.ukp.tu-darmstadt.de/data/argumentation-mining}
\end{abstract}

\section{Introduction}

Argumentation is a verbal activity which aims at increasing or decreasing the acceptability of a controversial standpoint \cite[p.~5]{vanEemeren1996}. 
It is a routine which is omnipresent in our daily verbal communication and thinking. 
Well-reasoned arguments are not only important for decision making and learning but also play a crucial role in drawing widely-accepted conclusions. 

\emph{Computational argumentation} is a recent research field in computational linguistics that focuses on the analysis of arguments in natural language texts. 
Novel methods have broad application potential in various areas like legal decision support \cite{MochalesPalau2009}, information retrieval \cite{Carstens2015}, policy making \cite{Sardianos2015}, and debating technologies \cite{Levy2014,Rinott2015}. 
Recently, computational argumentation has been receiving increased attention in \emph{computer-assisted writing} \cite{Song2014,Stab2014b} since it allows the creation of writing support systems that provide feedback about written arguments. 

Argumentation structures are closely related to discourse structures such as defined by \emph{rhetorical structure theory} (RST) \cite{Mann1987}, \emph{Penn discourse treebank} (PDTB) \cite{Prasad2008}, or \emph{segmented discourse representation theory} (SDRT) \cite{Asher2003}.
The internal structure of an \emph{argument} consists of several \emph{argument components}. 
It includes a \emph{claim} and one or more \emph{premises} \cite{Govier2010}.
The claim is a controversial statement and the central component of an argument, while premises are reasons for justifying (or refuting) the claim.
Moreover, arguments have directed \emph{argumentative relations}, describing the relationships one component has with another.
Each such relation indicates that the \emph{source component} is either a justification for or a refutation of the \emph{target component}. 

The identification of argumentation structures involves several subtasks like separating argumentative from non-argumentative text units \cite{Moens2007,Florou2013}, classifying argument components into claims and premises \cite{MochalesPalau2011,Rooney2012,Stab2014b}, and identifying argumentative relations \cite{MochalesPalau2009,Peldszus2014,Stab2014b}. 
However, an approach which covers all subtasks is still missing. 
Furthermore, most approaches operate locally and do not optimize the global argumentation structure. 
Recently, \namecite{Peldszus2015} proposed an approach based on \emph{minimum spanning trees} (MST) which jointly models argumentation structures. 
However, it links all argument components in a single tree structure.
Consequently, it is not capable of separating several arguments and recognizing unlinked argument components (e.g. unsupported claims).
In addition to the lack of end-to-end approaches for parsing argumentation structures, there are relatively few corpora annotated with argumentation structures at the discourse-level. 
Apart from our previous corpus \cite{Stab2014}, the few existing corpora lack non-argumentative text units \cite{Peldszus2014}, contain text genres different from our target domain \cite{Kirschner2015}, or the reliability is unknown \cite{Reed2008}.

Our primary motivation for this work is to create argument analysis methods for argumentative writing support systems and to achieve a better understanding of argumentation structures. 
Therefore, our first research question is whether human annotators can reliably identify argumentation structures in persuasive essays and if it is possible to create annotated data of high quality.
The second research question addresses the automatic recognition of argumentation structure. 
We investigate if, and how accurately, argumentation structures can be identified by computational techniques.
The contributions of this article are the following:
\begin{itemize}
\item \emph{An annotation scheme for modeling argumentation structures} derived from argumentation theory. Our annotation scheme models the argumentation structure of a document as a connected tree.
\item \emph{A novel corpus of 402 persuasive essays} annotated with discourse-level argumentation structures. We show that human annotators can apply our annotation scheme to persuasive essays with substantial agreement.
\item \emph{An end-to-end argumentation structure parser} which identifies argument components at the token level and globally optimizes component types and argumentative relations.
\end{itemize}
The remainder of this article is structured as follows: 
In Section \ref{relatedWork}, we review related work in computational argumentation and discuss the difference to traditional discourse analysis. 
In Section \ref{argStructures}, we derive our annotation scheme from argumentation theory. 
Section \ref{corpusCreation} presents the results of an annotation study and the corpus creation.
In Section \ref{approach}, we introduce the argumentation structure parser. 
We show that our model considerably improves the performance of base classifiers and significantly outperforms challenging heuristic baselines.
We conclude the article with a discussion in Section \ref{discussion}.

\section{Related Work}
\label{relatedWork}

Existing work in computational argumentation addresses a variety of different tasks. 
These include, for example, approaches for identifying reasoning type \cite{Feng2011}, argumentation style \cite{Oraby2015}, the stance of the author \cite{Somasundaran2009,Hasan2014}, the acceptability of arguments \cite{Cabrio2012a}, and appropriate support types \cite{Park2014}. 
Most relevant to our work, however, are approaches on \emph{argument mining} that focus on the identification of argumentation structures in natural language texts. 
We categorize related approaches into the following three subtasks:
\begin{itemize}
  \item \emph{Component identification} focuses on the separation of argumentative from non-argumentative text units and the identification of argument component boundaries. 
  \item \emph{Component classification} addresses the function of argument components. It aims at classifying argument components into different types such as claims and premises. 
  \item \emph{Structure identification} focuses on linking arguments or argument components. Its objective is to recognize different types of argumentative relations such as support or attack relations.
\end{itemize}

\subsection{Component Identification} 

\namecite{Moens2007} identified argumentative sentences in various types of text such as newspapers, parliamentary records and online discussions. 
They experimented with various different features and achieved an accuracy of $.738$ with word pairs, text statistics, verbs and keyword features.
\namecite{Florou2013} classified text segments as argumentative or non-argumentative using discourse markers and several features extracted from the tense and mood of verbs. They report an F1 score of $.764$. 
\namecite{Levy2014} proposed a pipeline including three consecutive steps for identifying context-dependent claims in Wikipedia articles. 
Their first component detects topic-relevant sentences including a claim. 
The second component detects the boundaries of each claim.
The third component ranks the identified claims for identifying the most relevant claims for the given topic.
\namecite{Goudas2014} presented a two-step approach for identifying argument components and their boundaries in social media texts. 
First, they classified each sentence as argumentative or non-argumentative and achieved an accuracy of $.774$. 
Second, they segmented each argumentative sentence using a \emph{conditional random field} (CRF). 
Their best model achieved an accuracy of $.424$.

\subsection{Component Classification}

The objective of the component classification task is to identify the type of argument components.
\namecite{Kwon2007} proposed two consecutive steps for identifying different types of claims in online comments. 
First, they classified sentences as claims and obtained an F1 score of $.55$ with a boosting algorithm.
Second, they classified each claim as either support, oppose or propose. 
Their best model achieved an F1 score of $.67$. 
\namecite{Rooney2012} applied kernel methods for classifying text units as either claims, premises or non-argumentative. 
They obtained an accuracy of $.65$. 
\namecite{MochalesPalau2011} classified sentences in legal decisions as claim or premise. 
They achieved an F1 score of $.741$ for claims and $.681$ for premises using a \emph{support vector machine} (SVM) with domain-dependent key phrases, text statistics, verbs, and the tense of the sentence. 
In our previous work, we used a multiclass SVM for labeling text units of student essays as major claim, claim, premise, or non-argumentative \cite{Stab2014b}. 
We obtained an accuracy of $.773$ using structural, lexical, syntactic, indicator and contextual features. 
Recently, \namecite{Nguyen2015} found that argument and domain words from unlabeled data increases accuracy to $.79$ using the same corpus, and \namecite{Lippi2015} achieved promising results using partial tree kernels for identifying sentences containing a claim. 

\subsection{Structure Identification}
\label{sec:sotaStructure}

Approaches on structure identification can be divided into macro-level approaches and micro-level approaches. 
Macro-level approaches such as presented by \namecite{Cabrio2012a}, \namecite{Ghosh2014}, or \namecite{Boltuvzic2014} address relations between complete arguments and ignore the microstructure of arguments. 
More relevant to our work, however, are micro-level approaches, which focus on relations between argument components. 
\namecite{MochalesPalau2009} introduced one of the first approaches for identifying the microstructure of arguments. 
Their approach is based on a manually created \emph{context-free grammar} (CFG) and recognizes argument structures as trees. 
However, it is tailored to legal argumentation and does not recognize \emph{implicit argumentative relations}, i.e. relations which are not indicated by discourse markers.
In previous work, we defined the task as the binary classification of ordered argument component pairs \cite{Stab2014b}. 
We classified each pair as support or not-linked using an SVM with structural, lexical, syntactic and indicator features. 
Our best model achieved an F1 score of $.722$. 
However, the approach recognizes argumentative relations locally and does not consider contextual information.
\namecite{Peldszus2014} modeled the targets of argumentative relations along with additional information in a single tagset. 
His tagset includes, for instance, several labels denoting if an argument component at position $n$ is argumentatively related to preceding argument components $n-1, n-2$, etc. or following argument components $n+1$, $n+2$, etc. 
Although his approach achieved a promising accuracy of $.48$, it is only applicable to short texts. 
\namecite{Peldszus2015} presented the first approach which globally optimizes argumentative relations. 
They jointly modeled several aspects of argumentation structures using an MST model and achieved an F1 score of $.720$.
They found that the function (support or attack) and the role (opponent and proponent) of argument components are the most useful dimensions for improving the identification of argumentative relations. 
Their corpus, however, is artificially created and includes a comparatively high proportion of opposing argument components (cf. Section \ref{existingCorpora}). 
Therefore, it is unclear whether the results can be reproduced with real data. 
Moreover, their approach links all argument components in a single tree structure. 
Thus, it is not capable of separating several arguments and recognizing unlinked components.

\subsection{Existing Corpora Annotated with Argumentation Structures} 
\label{existingCorpora}

Existing corpora in computational argumentation cover numerous aspects of argumentation analysis. 
There are, for instance, corpora which address argumentation strength \cite{Persing2015}, factual knowledge \cite{Klebanov2012}, various properties of arguments \cite{Walker2012}, argumentative relations between complete arguments at the macro-level  \cite{Cabrio2014,Boltuvzic2014}, different types of argument components \cite{MochalesPalau2009b,Kwon2007,Habernal2016}, and argumentation structures over several documents \cite{Aharoni2014}.
However, corpora annotated with argumentation structures at the level of discourse are still rare.

One prominent resource is AraucariaDB \cite{Reed2008}. 
It includes heterogenous text types such as newspaper editorials, parliamentary records, judicial summaries and online discussions. 
It also includes annotations for the reasoning type and implicit argument components, which were added by the annotators during the analysis.
However, the reliability of the annotations is unknown.

\namecite{Kirschner2015} annotated argumentation structures in introduction and discussion sections of $24$ German scientific articles. 
Their annotation scheme includes four argumentative relations (support, attack, detail and sequence). 
However, the corpus does not include annotations for argument component types. 

\namecite{Peldszus2015} created a small corpus of $112$ German microtexts with controlled linguistic and rhetoric complexity. 
Each document includes a single argument and does not include more than five argument components.
Their annotation scheme models supporting and attacking relations as well as additional information like proponent and opponent. 
They obtained an \emph{inter-annotator agreement} (IAA) of $\kappa=.83$ with three expert annotators. 
Recently, they translated the corpus to English resulting in the first parallel corpus for computational argumentation.
However, the corpus does not include non-argumentative text units. 
Therefore, the corpus is only of limited use for training end-to-end argumentation structure parsers.
Due to the employed writing guidelines \cite[p.~197]{Peldszus2013}, it also exhibits an unusually high proportion of attack relations. 
In particular, $97$ of the $112$ arguments (86.6\%) include at least one attack relation.

\begin{table}[!ht]
\caption{Existing corpora annotated with argumentation structures at the discourse-level (\#Doc = number of documents; \#Comp = number of argument components; NoArg = presence of non-argumen- tative text units; *Recent releases do not include non-argumentative text units).}\label{tab:corpora}
\footnotesize{
	\begin{tabularx}{\textwidth}{  p{0.25\textwidth}  | p{0.16\textwidth} | p{0.05\textwidth} | p{0.07\textwidth} | p{0.06\textwidth} | p{0.10\textwidth} | p{0.09\textwidth}  }

 \centering{\emph{Source}} & \centering{\emph{Genre}} & \centering{\emph{\#Doc}}  & \centering{\emph{\#Comp}} & \centering{\emph{NoArg}} & \centering{\emph{Granularity}} & \centering{\emph{IAA}} \tabularnewline
\hline
\centering{\cite{Reed2008}} & \centering{various} & \centering{\textasciitilde$700$} & \centering{\textasciitilde$2{,}000$} & \centering{yes*} & \centering{clause} & \centering{unknown} \tabularnewline
\centering{\cite{Stab2014}} & \centering{student essays} & \centering{$90$} & \centering{$1{,}552$} & \centering{yes} &  \centering{clause} & \centering{$\alpha_U = .72$} \tabularnewline
\centering{\cite{Peldszus2015}} & \centering{microtexts} & \centering{$112$} & \centering{$576$} & \centering{no} &  \centering{clause} & \centering{$\kappa = .83$} \tabularnewline
\centering{(Kirschner et al. 2015)} & \centering{scientific articles} & \centering{$24$} & \centering{\textasciitilde$2{,}700$} & \centering{yes} &  \centering{sentence} & \centering{$\kappa = .43$} \tabularnewline
	\end{tabularx}
	}
\end{table}

In previous work, we created a corpus of $90$ persuasive essays, which we  selected randomly from essayforum.com \cite{Stab2014}. We annotated the corpus in two consecutive steps:  
First, we identified argument components at the clause level and obtained an inter-annotator agreement of $\alpha_U = .72$ between three annotators. 
Second, we annotated argumentative support and attack relations between argument  components and achieved an inter-annotator agreement of $\kappa = .8$. 
In contrast to the microtext corpus from Peldszus, the corpus includes non-argumentative text units and exhibits a more realistic proportion of argumentative attack relations since the essays were not written in a controlled experiment. 
Apart from this corpus, we are only aware of one additional study on argumentation structures in persuasive essays. 
\namecite{Botley2014} analyzed $10$ essays using argument diagramming for studying differences in argumentation strategies. 
Unfortunately, the corpus is too small for computational purposes and the reliability of the annotations is unknown. 
Table \ref{tab:corpora} provides an overview of existing corpora annotated with argumentation structures at the discourse-level.

\subsection{Discourse Analysis}

The identification of argumentation structures is closely related to discourse analysis. 
Similar to the identification of argumentation structures, discourse analysis aims at identifying elementary discourse units and discourse relations between them. 
Existing approaches on discourse analysis mainly differ in the employed discourse theory. 
RST \cite{Mann1987}, for instance, models discourse structures as trees by iteratively linking adjacent discourse units \cite{Feng2014,Hernault2010} while approaches based on PDTB \cite{Prasad2008} identify more shallow structures by linking two adjacent sentences or clauses \cite{Lin2014}. 
Whereas RST and PDTB are limited to discourse relations between adjacent discourse units, SDRT \cite{Asher2003} also allows long distance relations \cite{Afantenos2014,Afantenos2015}. 
However, similar to argumentation structure parsing the main challenge of discourse analysis is to identify implicit discourse relations \cite[p.~1694]{Braud2014}.

\namecite{Marcu2002} proposed one of the first approaches for identifying implicit discourse relations. 
In order to collect large amounts of training data, they exploited several discourse markers like ``because'' or ``but''. 
After removing the discourse markers, they found that word pair features are useful for identifying implicit discourse relations. 
\namecite{Pitler2009} proposed an approach for identifying four implicit types of discourse relations in the PDTB and achieved F1 scores between $.22$ and $.76$. 
They found that using features tailored to each individual relation leads to the best results. 
\namecite{Lin2009} showed that production rules collected from parse trees yield good results and \namecite{Louis2010} found that features based on named entities do not perform as well as lexical features. 

Approaches to discourse analysis usually aim at identifying various different types of discourse relations.
However, only a subset of these relations is relevant for argumentation structure parsing. 
For example, \namecite{Peldszus2013} proposed support, attack and counter-attack relations for modeling argumentation structures, whereas our work focuses on support and attack relations. 
This difference is also illustrated by the work of \namecite{Biran2011}. 
They selected a subset of $12$ relations from the \emph{RST discourse treebank} \cite{Carlson2001} and argue that only a subset of RST relations is relevant for identifying justifications.

\section{Argumentation: Theoretical Background}
\label{argStructures}

The study of argumentation is a comprehensive and interdisciplinary research field. 
It involves philosophy, communication science, logic, linguistics, psychology, and computer science. 
The first approaches to studying argumentation date back to the ancient Greek sophists and evolved in the 6$^{th}$ and 5$^{th}$ centuries B.C. \cite{vanEemeren1996}.
In particular, the influential works of Aristotle on traditional logic, rhetoric, and dialectics set an important milestone and are a cornerstone of modern argumentation theory.
Due to the diversity of the field, there are numerous proposals for modeling argumentation. 
\namecite{Bentahar2010} categorize argumentation models into three types: (\emph{i}) monological models, (\emph{ii}) dialogical models, and \emph{(iii)} rhetorical models.
\emph{Monological models} address the internal microstructure of arguments. 
They focus on the function of argument components, the links between them, and the reasoning type. 
Most monological models stem from the field of informal logic and focus on arguments as product \cite{Johnson2000,OKeefe1977}. 
On the other hand, \emph{dialogical models} focus on the process of argumentation and ignore the  microstructure of arguments. 
They model the external macrostructure and address relations between arguments in dialogical communications.
Finally, \emph{rhetorical models} consider neither the micro- nor the macrostructure but rather the way arguments are used as a means of persuasion. 
They consider the audience's perception and aim at studying rhetorical schemes that are successful in practice. 
In this article, we focus on the monological perspective which is well-suited for developing computational methods \cite{Peldszus2013,Lippi2016}.

\subsection{Argument Diagramming}

The laying out of argument structure is a widely used method in informal logic \cite{Copi1990,Govier2010}. 
This technique, referred to as \emph{argument diagramming}, aims at transferring natural language arguments into a structured representation for evaluating them in subsequent analysis steps \cite[p. 447]{Henkemans2000}. 
Although argumentation theorists consider argument diagramming a manual activity, the diagramming conventions also serve as a good foundation for developing novel argument mining models \cite{Peldszus2013}. 
\begin{figure}[h]
\centering
\includegraphics[width=1.0\linewidth]{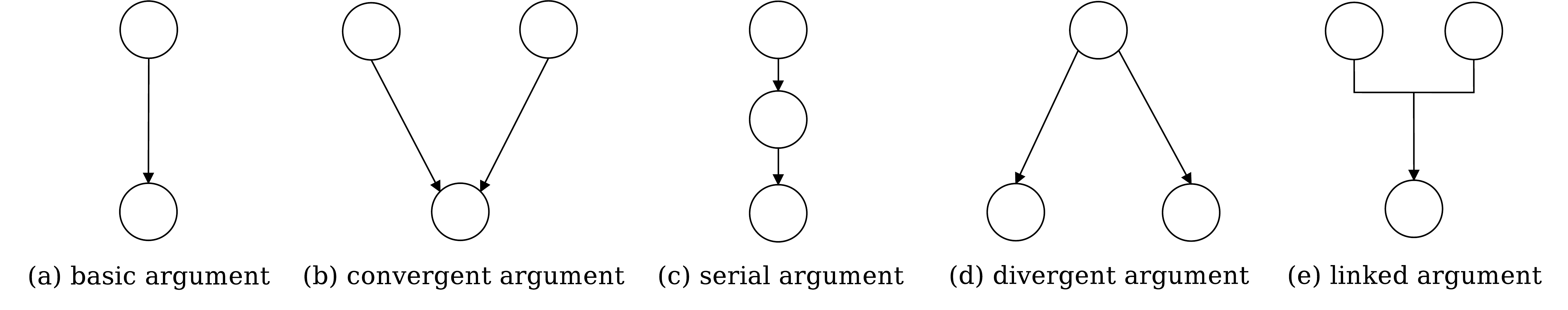}
\caption{Microstructures of arguments: nodes are argument components and links represent argumentative relations. Nodes at the bottom are the claims of the arguments.} 
\label{fig:ArgStructures}
\end{figure}
An argument diagram is a node-link diagram whereby each node represents an argument component, i.e. a statement represented in natural language and  
each link represents a directed argumentative relation indicating that the \emph{source component} is a justification (or refutation) of the \emph{target component}. 
For example, Figure \ref{fig:ArgStructures} shows some common argument structures. 
A \emph{basic argument} includes a claim supported by a single premise. 
It can be considered the minimal form an argument can take. 
A \emph{convergent argument} comprises two premises that support the claim individually; an argument is \emph{serial} if it includes a reasoning chain and \emph{divergent} if a single premise supports several claims \cite{Beardsley1950}. 
Complementarily, \namecite{Thomas1973} defined \emph{linked arguments} (Figure \ref{fig:ArgStructures}e). 
Like convergent arguments, a linked argument includes two premises. 
However, neither of the two premises independently supports the claim. 
The premises are only relevant to the claim in conjunction. 
More complex arguments can combine any of these elementary structures illustrated in Figure \ref{fig:ArgStructures}. 

On closer inspection, however, there are several ambiguities when applying argument diagramming to real texts: 
First, the distinction between convergent and linked structures is often ambiguous in real argumentation structures \cite{Henkemans2000,Freeman2011}. 
Second, it is unclear if the argumentation structure is a graph or a tree. 
Third, the argumentative type of argument components is ambiguous in serial structures. 
We discuss each of these questions in the following sections.

\subsubsection{Distinguishing between Linked and Convergent Arguments}

The question if an argumentation model needs to distinguish between linked and convergent arguments is still debated in argumentation theory \cite{vanEemeren1996,Freeman2011,Yanal1991,Conway1991}. 
From a perspective based on traditional logic, linked arguments indicate \emph{deductive reasoning} and convergent arguments represent \emph{inductive reasoning} \cite[p.~453]{Henkemans2000}. 
However, \namecite[p.~91ff.]{Freeman2011} showed that the traditional definition of linked arguments is frequently ambiguous in everyday discourse.
\namecite{Yanal1991} argues that the distinction is equivalent to separating several arguments and \namecite{Conway1991} argues that linked structures can simply be omitted for modeling single arguments. 
From a computational perspective, the identification of linked arguments is equivalent to finding groups of premises or classifying the reasoning type of an argument as either deductive or inductive. 
Accordingly, it is not necessary to distinguish linked and convergent arguments during the identification of argumentation structures since this task can be solved in subsequent analysis steps.

\subsubsection{Argumentation Structures as Trees} 
\label{argStructureAsTree}

Defining argumentation structures as trees implies the exclusion of divergent arguments, to allow only one target for each premise and to neglect cycles. 
From a theoretical perspective, divergent structures are equivalent to several arguments (one for each claim) \cite[p.~16]{Freeman2011}. 
As a result of this treatment, a great many of theoretical textbooks neglect divergent structures \cite{Henkemans2000,Reed2004} and also most computational approaches consider arguments as trees \cite{MochalesPalau2009,Cohen1987,Peldszus2014}.
However, there is little empirical evidence regarding the structure of arguments. 
We are only aware of one study which showed that 5.26\% of the arguments in political speeches (which can be assumed to exhibit complex argumentation structures) are divergent.

Essay writing usually follows a ``claim-oriented'' procedure \cite{Whitaker2009,Shiach2009,Perutz2010,Kemper2004}. 
Starting with the formulation of the standpoint on the topic, authors collect claims in support (or opposition) of their view. 
Subsequently, they collect premises that support or attack their claims. 
The following example illustrates this procedure. 
A major claim on abortion, for instance, is ``\emph{abortion should be illegal}''; a supporting claim could be ``\emph{abortion is ethically wrong}'' and the associated premises ``\emph{unborn babies are considered human beings}'' and ``\emph{killing human beings is wrong}''.
Due to this common writing procedure, divergent and circular structures are rather unlikely in persuasive essays. 
Therefore, we assume that modeling the argumentation structure of essays as a tree is a reasonable decision.

\subsubsection{Argumentation Structures and Argument Component Types} 
\label{sec:argStructuresAndTypes}
Assigning argumentative types to the components of an argument is unambiguous if the argumentation structure is shallow. 
It is, for instance, obvious that an argument component $c_1$ is a premise and argument component $c_2$ is a claim, if $c_1$ supports $c_2$ in a basic argument (cf. Figure \ref{fig:ArgStructures}). 
However, if the tree structure is deeper, i.e. exhibits serial structures, assigning argumentative types becomes ambiguous. 
Essentially, there are three different approaches for assigning argumentative types to argument components. 
First, according to \namecite{Beardsley1950} a serial argument includes one argument component which is both a claim and a premise. 
Therefore, the inner argument component bears two different argumentative types (\emph{multi-label approach}). 
Second, \namecite[p.~24]{Govier2010} distinguishes between ``main claim'' and ``subclaim''. Similarly, \namecite[p.~17]{Damer2009} distinguishes between ``premise'' and ``subpremise'' for labeling argument components in serial structures. 
Both approaches define specific labels for each level in the argumentation structure (\emph{level approach}). 
Third, \namecite{Cohen1987} considers only the root node of an argumentation tree as a claim and the following nodes in the structure as premises (\emph{``one-claim'' approach}). 
In order to define an argumentation model for persuasive essays, we propose a \emph{hybrid approach} that combines the level approach and the ``one-claim'' approach.

\subsection{Argumentation Structures in Persuasive Essays}
\label{annotationScheme}

We model the argumentation structure of persuasive essays as a connected tree structure. 
We use a level approach for modeling the first level of the tree and a ``one-claim'' approach for representing the structure of each individual argument. 
Accordingly, we model the first level of the tree with two different argument component types and the structure of individual arguments with argumentative relations. 

The \emph{major claim} is the root node of the argumentation structure and represents the author's standpoint on the topic. 
It is an opinionated statement that is usually stated in the introduction and restated in the conclusion of the essay. 
The individual body paragraphs of an essay include the actual arguments. 
They either support or attack the author's standpoint expressed in the major claim. 
Each argument consists of a claim and several premises. 
In order to differentiate between supporting and attacking arguments, each claim has a \emph{stance attribute} that can take the values ``for'' or ``against''. 

We model the structure of each argument with a ``one-claim'' approach. 
The \emph{claim} constitutes the central component of each argument. 
The \emph{premises} are the reasons of the argument. 
The actual structure of an argument comprises directed argumentative support and attack relations, which link a premise either to a claim or to another premise (serial arguments). 
Each premise $p$ has one \emph{outgoing relation}, i.e. there is a relation that has $p$ as source component, and none or several \emph{incoming relations}, i.e. there can be a relation with $p$ as target component.
A claim can exhibit several incoming relations but no outgoing relation.
The ambiguous function of inner premises in serial arguments is implicitly modeled by the structure of the argument. 
The inner premise exhibits one outgoing relation and at least one incoming relation. 
Finally, the stance of each premise is indicated by the type of its outgoing relation (support or attack). 

\begin{figure}[h]
\centering
\includegraphics[width=1.0\linewidth]{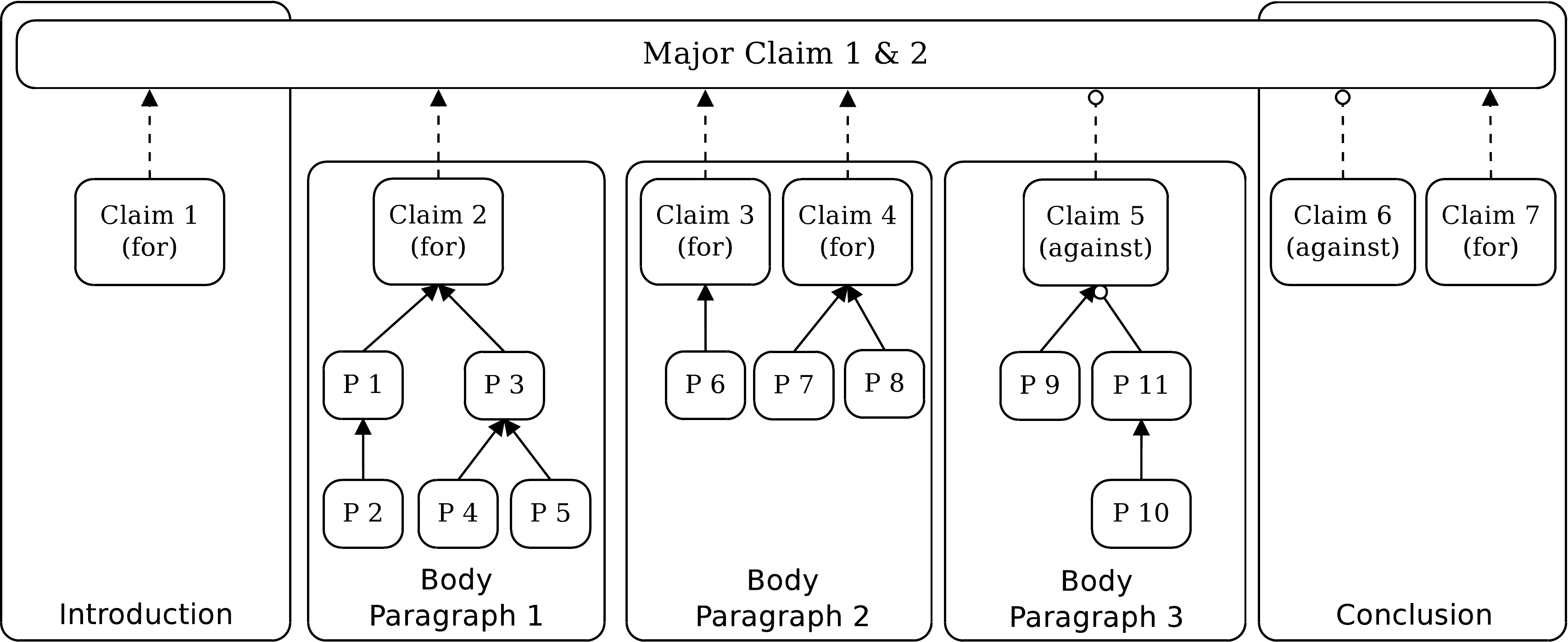}
\caption{Argumentation structure of the example essay. Arrows indicate argumentative relations. Arrowheads denote argumentative support relations and circleheads attack relations. Dashed lines indicate relations that are encoded in the stance attributes of claims. ``P'' denotes premises.} 
\label{fig:essayArgStructure}
\end{figure}

The following example illustrates the argumentation structure of a persuasive essay.\footnote{The example essay was written by the authors to illustrate all phenomena of argumentation structures in persuasive essays.} The introduction of an essay describes the controversial topic and usually includes the major claim:

\begin{quote}
\emph{Ever since researchers at the Roslin Institute in Edinburgh cloned an adult sheep, there has been an ongoing debate about whether cloning technology is morally and ethically right or not. Some people argue for and others against and there is still no agreement whether cloning technology should be permitted. However, as far as I'm concerned, [\textbf{cloning is an important technology for humankind}]$_{MajorClaim1}$ since [\ul{it would be very useful for developing novel cures}]$_{Claim1}$.}
\end{quote}
\noindent
The first two sentences introduce the topic and do not include argumentative content. The third sentence contains the major claim (boldfaced) and a claim which supports the major claim (underlined). The following body paragraphs of the essay include arguments which either support or attack the major claim. For example, the following body paragraph includes one argument that supports the positive standpoint of the author on cloning: 

\begin{quote}
\emph{First, [\ul{cloning will be beneficial for many people who are in need of organ transplants}]$_{Claim2}$. [\uwave{Cloned organs will match perfectly to the blood group and tissue of patients}]$_{Premise1}$ since [\uwave{they can be raised from cloned stem cells of the patient}]$_{Premise2}$. In addition, [\uwave{it shortens the healing process}]$_{Premise3}$. Usually, [\uwave{it is very rare to find an appropriate organ donor}]$_{Premise4}$ and [\uwave{by using cloning in order to raise required organs the waiting time can be shortened tremendously}]$_{Premise5}$.}
\end{quote}
\noindent
The first sentence contains the claim of the argument, which is supported by five premises in the following three sentences (wavy underlined). The second sentence includes two premises, of which premise$_1$ supports claim$_2$ and premises$_2$ supports premise$_1$. Premise$_3$ in the third sentence supports claim$_2$. The fourth sentence includes premise$_4$ and premise$_5$. Both support premise$_3$. The next paragraph illustrates a body paragraph with two arguments:

\begin{quote}
\emph{Second, [\uwave{scientists use animals as models in order to learn about human diseases}]$_{Premise6}$ and therefore [\ul{cloning animals enables novel developments in science}]$_{Claim3}$. Furthermore, [\uwave{infertile couples can have children that are genetically related}]$_{Premise7}$. [\uwave{Even same sex couples can have children}]$_{Premise8}$. Consequently, [\ul{cloning can help families to get children}]$_{Claim4}$.}
\end{quote}
\noindent
The initial sentence includes the first argument, which consists of premise$_6$ and claim$_3$. 
The following three sentences include the second argument. 
Premise$_7$ and premise$_8$ both support claim$_4$ in the last sentence.
Both arguments cover different aspects (development in science and cloning humans) which both support the author's standpoint on cloning. 
This example illustrates that knowing argumentative relations is important for separating several arguments in a paragraph. 
The example also shows that argument components frequently exhibit preceding text units that are not relevant to the argument but helpful for recognizing the argument component type. 
For example, preceding discourse connectors like ``therefore'', ``consequently'', or ``thus'' can signal a subsequent claim. 
Discourse markers like ``because'', ``since'', or ``furthermore'' could indicate a premises. 
We refer to these text units as \emph{preceding tokens}. 
The third body paragraph illustrates a contra argument and argumentative attack relations:

\begin{quote}
\emph{Admittedly, [\ul{cloning could be misused for military purposes}]$_{Claim5}$. For example, [\uwave{it could be used to manipulate human genes in order to create obedient soldiers with extraordinary abilities}]$_{Premise9}$. However, because [\uwave{moral and ethical values are internationally shared}]$_{Premise10}$, [\uwave{it is very unlikely that cloning will be misused for militant objectives}]$_{Premise11}$. }
\end{quote}
\noindent
The paragraph begins with claim$_5$, which attacks the stance of the author. It is supported by premise$_9$ in the second sentence. 
The third sentence includes two premises, both of which defend the stance of the author.  
Premise$_{11}$ is an attack of claim$_5$ and premise$_{10}$ supports premise$_{11}$.
The last paragraph (conclusion) restates the major claim and summarizes the main aspects of the essay:

\begin{quote}
\emph{To sum up, although [\ul{permitting cloning might bear some risks like misuse for military purposes}]$_{Claim6}$, I strongly believe that [\textbf{this technology is beneficial to humanity}]$_{MajorClaim2}$. It is likely that [\ul{this technology bears some important cures which will significantly improve life conditions}]$_{Claim7}$.}
\end{quote}
\noindent
The conclusion of the essay starts with an attacking claim followed by the restatement of the major claim. The last sentence includes another claim that summarizes the most important points of the author's argumentation. Figure \ref{fig:essayArgStructure} shows the entire argumentation structure of the example essay. 

\section{Corpus Creation}
\label{corpusCreation}

The motivation for creating a new corpus is threefold: 
First, our previous corpus is relatively small. 
We believe that more data will improve the accuracy of our computational models.
Second, we ensure the reproducibility of the annotation study and validate our previous results. 
Third, we improved our annotation guidelines.
We added more precise rules for segmenting argument components and a detailed description of common essay structures. 
We expect that our novel annotation guidelines will guide annotators towards  adequate agreement without collaborative training sessions. 
Our annotation guidelines comprise 31 pages and include the following three steps: 

\begin{enumerate}
\item \emph{Topic and stance identification}: We found in our previous annotation study that knowing the topic and stance of an essay improves inter-annotator agreement \cite{Stab2014}. For this reason, we ask the annotators to read the entire essay before starting with the annotation task.
\item \emph{Annotation of argument components}: Annotators mark major claims, claims and premises. They annotate the boundaries of argument components and determine the stance attribute of claims.
\item \emph{Linking premises with argumentative relations}: The annotators identify the structure of arguments by linking each premise to a claim or another premise with argumentative support or attack relations.
\end{enumerate}

\noindent
Three non-native speakers with excellent English proficiency participated in our annotation study. 
One of the three annotators already participated in our previous study (expert annotator). 
The two other annotators learned the task by independently reading the annotation guidelines. 
We used the \emph{brat rapid annotation tool} \cite{Stenetorp2012}.
It provides a graphical web interface for marking text units and linking them.

\subsection{Data}

We randomly selected $402$ English essays from essayforum.com. 
This online forum is an active community which provides correction and feedback about different texts such as research papers, essays, or poetry. 
For example, students post their essays in order to receive feedback about their writing skills while preparing for standardized language tests.
We manually reviewed each essay and selected only those with a sufficiently detailed description of the writing prompt. 
The corpus includes $7{,}116$ sentences with $147{,}271$ tokens.

\subsection{Inter-Annotator Agreement}
\label{sec:IAA}

All three annotators independently annotated a random subset of $80$ essays. 
The remaining $322$ essays were annotated by the expert annotator. 
We evaluate the inter-annotator agreement of the argument component annotations using two different strategies:
First, we evaluate if the annotators agree on the presence of argument components in sentences using \emph{observed agreement} and \emph{Fleiss'~$\kappa$} \cite{Fleiss1971}. 
We consider each sentence as a markable and evaluate the presence of each argument component type $t\in\{MajorClaim,Claim,Premise\}$ in a sentence individually. 
Accordingly, the number of markables for each argument component type $t$ corresponds to the number of sentences $N=1{,}441$, the number of annotations per markable equals with the number of annotators $n=3$, and the number of categories is $k=2$ (``$t$'' or ``\emph{not $t$}'').
Evaluating the agreement at the sentence level is an approximation of the actual agreement since the boundaries of argument components can differ from sentence boundaries and a sentence can include several argument components.\footnote{In our evaluation set of $80$ essays the annotators identified in $4.3\%$ of the sentences several argument components of different types. Thus, evaluating the reliability of argument components at the sentence level is a good approximation of the inter-annotator agreement.}
Therefore, for the second evaluation strategy, we employ Krippendorff's $\alpha_U$ \cite{Krippendorff2004} which considers the differences in the component boundaries at the token level. 
Thus, it allows for assessing the reliability of our annotation study more accurately. 
For determining the inter-annotator agreement, we use \emph{DKPro Agreement} whose implementations of inter-annotator agreement measures are well-tested with various examples from literature \cite{Meyer2014}.

\begin{table}[!ht]
\caption{Inter-annotator agreement of argument components.}\label{tab:iaaComponents}
\footnotesize{
	\begin{tabularx}{\textwidth}{p{0.22\textwidth} | p{0.22\textwidth} | p{0.22\textwidth} | p{0.22\textwidth}}
		 \centering{\emph{Component type}} & \centering{\emph{Observed agreement}} & \centering{\emph{Fleiss'~$\kappa$}} & \centering{\emph{$\alpha_U$}}\tabularnewline
		\hline
		\centering{MajorClaim} & \centering{$97.9\%$} & \centering{$.877$} &  \centering{$.810$} \tabularnewline
		\centering{Claim} &  \centering{$88.9\%$} &  \centering{$.635$} &  \centering{$.524$} \tabularnewline
		\centering{Premise}& \centering{$91.6\%$} & \centering{$.833$} & \centering{$.824$} \tabularnewline
	\end{tabularx}
	}
\end{table} 

Table \ref{tab:iaaComponents} shows the inter-annotator agreement of each argument component type. 
The agreement is best for major claims.
The IAA scores of $97.9\%$ and $\kappa = .877$ indicate that annotators reliably identify major claims in persuasive essays. 
In addition, the unitized alpha measure of $\alpha_U = .810$ shows that there are only few disagreements about the boundaries of major claims. 
The results also indicate good agreement for premises ($\kappa = .833$ and $\alpha_U = .824$). 
We obtain the lowest agreement of $\kappa = .635$ for claims which shows that the identification of claims is more complex than identifying major claims and premises. 
The joint unitized measure for all argument components is $\alpha_U = .767$, and thus the agreement improved by $.043$ compared to our previous study \cite{Stab2014b}. 
Therefore, we conclude that human annotators can reliably annotate argument components in persuasive essays.

For determining the agreement of the stance attribute, we follow the same methodology as for the sentence level agreement described above, but we consider each sentence containing a claim as ``for'' or ``against'' according to its stance attribute and all sentences without a claim as ``none''.
Consequently, the agreement of claims constitutes the upper bound for the stance attribute. 
We obtain an agreement of $88.5\%$ and $\kappa = .623$ which is slightly below the agreement scores of claims (cf. Table \ref{tab:iaaComponents}). 
Therefore, human annotators can reliably differentiate between supporting and attacking claims.

We determined the markables for evaluating the agreement of argumentative relations by pairing all argument components in the same paragraph. 
For each paragraph with argument components $c_1,...,c_n$, we consider each pair $p=(c_i,c_j)$ with $1 \leq i,j \leq n$ and $i\neq j$ as markable. 
Thus, the set of all markables corresponds to all argument component pairs that can be annotated according to our guidelines.
The number of argument component pairs is $N=4{,}922$, the number of ratings per markable is $n=3$, and the number of categories $k=2$.
\begin{table}[!ht]
\caption{Inter-annotator of argumentative relations.}\label{tab:iaaRelations}
\footnotesize{
	\begin{tabularx}{\textwidth}{p{0.29\textwidth}  | p{0.29\textwidth} | p{0.29\textwidth} }
		 \centering{\emph{Relation type}} & \centering{\emph{Observed agreement}} & \centering{\emph{Fleiss' $\kappa$}} \tabularnewline
		\hline
		\centering{Support} & \centering{$.923$} & \centering{$.708$}  \tabularnewline
		\centering{Attack} &  \centering{$.996$} & \centering{$.737$} \tabularnewline
	\end{tabularx}
	}
\end{table} 

Table \ref{tab:iaaRelations} shows the inter-annotator agreement of argumentative relations. 
We obtain for both argumentative support and attack relations $\kappa$-scores above $.7$ which allows tentative conclusions \cite{Krippendorff2004}. 
On average the annotators marked only $0.9\%$ of the $4{,}922$ pairs as argumentative attack relations and $18.4\%$ as argumentative support relations.
Although the agreement is usually much lower if a category is rare \cite[p.~573]{Artstein2008}, the annotators agree more on argumentative attack relations. 
This indicates that the identification of argumentative attack relations is a simpler task than identifying argumentative support relations. 
The agreement scores for argumentative relations are approximately $.10$ lower compared to our previous study. 
This difference can be attributed to the fact that we did not explicitly annotate relations between claims and major claims which are easy to annotate due to the known types of argument components (cf. Section \ref{annotationScheme}).

\subsection{Analysis of Human Disagreement}

For analyzing the disagreements between the annotators, we determined \emph{confusion probability matrices} (CPM) \cite{Cinkova2012}. 
Compared to traditional confusion matrices, a CPM also allows to analyze confusion if more than two annotators are involved in an annotation study. 
A CPM includes conditional probabilities that an annotator assigns a category in the column given that another annotator selected the category in the row. 
\begin{table}[!ht]
\caption{Confusion probability matrix of argument component annotations (``NoArg'' indicates sentences without argumentative content).}
\label{tab:cpmComponents}
\footnotesize{
	\begin{tabularx}{\textwidth} {  p{0.16\textwidth}  | p{0.16\textwidth} | p{0.16\textwidth} | p{0.16\textwidth} | p{0.16\textwidth} }
		 & \centering{\emph{MajorClaim}} & \centering{\emph{Claim}} & \centering{\emph{Premise}} & \centering{\emph{NoArg}} \tabularnewline
		\hline
		\emph{MajorClaim} & \centering{$.771$} & \centering{$.077$} & \centering{$.010$} & \centering{$.142$}\tabularnewline
		\emph{Claim} &  \centering{$.036$} &  \centering{$.517$} &  \centering{$.307$} & \centering{$.141$}\tabularnewline
		\emph{Premise} &  \centering{$.002$} &  \centering{$.131$} &  \centering{$.841$} & \centering{$.026$}\tabularnewline
		\emph{NoArg} &  \centering{$.059$} &  \centering{$.126$} &  \centering{$.054$} & \centering{$.761$}\tabularnewline
	\end{tabularx}
	}
\end{table} 
Table \ref{tab:cpmComponents} shows the CPM of argument component annotations. 
It shows that the highest confusion is between claims and premises. 
We observed that one annotator frequently did not split sentences including a claim.
For instance, the annotator labeled the entire sentence as a claim although it includes an additional premise. 
This type of error also explains the lower unitized alpha score compared to the sentence level agreements in Table \ref{tab:iaaComponents}. 
Furthermore, we found that concessions before claims were frequently not annotated as an attacking premise. 
For example, annotators often did not split sentences similar to the following example:
\begin{quote}
\emph{Although [\uwave{in some cases technology makes people's life more complicated}]$_{premise}$, [\ul{the convenience of technology outweighs its drawbacks}]$_{claim}$.}
\end{quote}

The distinction between major claims and claims exhibits less confusion.
This may be due to the fact that major claims are relatively easy to locate in essays since they occur usually in introductions or conclusions whereas claims can occur anywhere in the essay. 

\begin{table}[!ht]
\caption{Confusion probability matrix of argumentative relation annotations (``Not-Linked'' indicates argument component pairs which are not argumentatively related).}
\label{tab:cpmRelations}
\footnotesize{
	\begin{tabularx}{\textwidth} {  p{0.22\textwidth}  | p{0.22\textwidth} | p{0.22\textwidth} | p{0.22\textwidth} }
		 & \centering{\emph{Support}} & \centering{\emph{Attack}} & \centering{\emph{Not-Linked}} \tabularnewline
		\hline
		\emph{Support} & \centering{$.605$} & \centering{$.006$} & \centering{$.389$} \tabularnewline
		\emph{Attack} &  \centering{$.107$} & \centering{$.587$} & \centering{$.307$} \tabularnewline
		\emph{Not-Linked} &  \centering{$.086$} & \centering{$.004$} & \centering{$.910$} \tabularnewline
	\end{tabularx}
}
\end{table} 

\noindent
Table \ref{tab:cpmRelations} shows the CPM of argumentative relations. 
There is little confusion between argumentative support and attack relations. 
The CPM also shows that the highest confusion is between argumentative relations (support and attack) and unlinked pairs.
This can be attributed to the identification of the correct targets of premises. 
In particular, we observed that agreement on the targets decreases if a paragraph includes several claims or serial argument structures.

\subsection{Creation of the Final Corpus}

We created a partial gold standard of the essays annotated by all annotators.
We use this partial gold standard of $80$ essays as our test data ($20\%$) and the remaining $322$ essays annotated by the expert annotator as our training data ($80\%$). 
The creation of our gold standard test data consists of the following two steps: 
first, we merge the annotation of all argument components. 
Thus, each annotator annotates argumentative relations based on the same argument components. 
Second, we merge the argumentative relations to compile our final gold standard  test data. 
Since the argument component types are strongly related - the selection of the premises, for instance, depends on the selected claim(s) in a paragraph - we did not merge the annotations using majority voting as in our previous study. 
Instead, we discussed the disagreements in several meetings with all annotators for resolving the disagreements.

\subsection{Corpus Statistics}
\label{sec:corpusStatistics}

Table \ref{tab:statistics} shows an overview of the size of the corpus. 
It contains $6{,}089$ argument components, $751$ major claims, $1{,}506$ claims, and $3{,}832$ premises. 
Such a large proportion of claims compared to premises is common in argumentative texts since writers tend to provide several reasons for ensuring a robust standpoint \cite{MochalesPalau2011}.

\begin{table}[!ht]
\caption{Statistics of the final corpus.}\label{tab:statistics}
\footnotesize{
	\begin{tabularx}{\textwidth}{  p{0.005\textwidth}  p{0.22\textwidth} | p{0.20\textwidth} | p{0.20\textwidth} | p{0.20\textwidth}  }

		&& \centering{\emph{all}} & \centering{\emph{avg. per essay}} & \centering{\emph{standard deviation}} \tabularnewline
		\hline
		
		\parbox[t]{2mm}{\multirow{3}{*}{\rotatebox[origin=c]{90}{\emph{size}}}} & Sentences & \centering{$7{,}116$} & \centering{$18$} & \centering{$4.2$} \tabularnewline
		&Tokens & \centering{$147{,}271$} & \centering{$366$} & \centering{$62.9$} \tabularnewline
		&Paragraphs & \centering{$1{,}833$} & \centering{$5$} & \centering{$0.6$} \tabularnewline
		
		\hline
		\parbox[t]{2mm}{\multirow{6}{*}{\rotatebox[origin=c]{90}{\emph{arg. comp.}}}} &Arg. components & \centering{$6{,}089$} & \centering{$15$} & \centering{$3.9$} \tabularnewline
		
		&MajorClaims & \centering{$751$} & \centering{$2$} & \centering{$0.5$} \tabularnewline
		&Claims & \centering{$1{,}506$} & \centering{$4$} & \centering{$1.2$} \tabularnewline

		&Premises & \centering{$3{,}832$} & \centering{$10$} & \centering{$3.4$} \tabularnewline

		& Claims (for) & \centering{$1{,}228$} & \centering{$3$} & \centering{$1.3$} \tabularnewline

		& Claims (against) & \centering{$278$} & \centering{$1$} & \centering{$0.8$} \tabularnewline 
		\hline 
		\parbox[t]{2mm}{\multirow{2}{*}{\rotatebox[origin=c]{90}{\emph{rel.}}}}
		& Support & \centering{$3{,}613$} & \centering{$9$} & \centering{$3.3$} \tabularnewline

		& Attack & \centering{$219$} & \centering{$1$} & \centering{$0.9$} \tabularnewline

	\end{tabularx}
	}
\end{table} 

The proportion of non-argumentative text amounts to 47,474 tokens (32.2\%) and 1,631 sentences (22.9\%).
The number of sentences with several argument components is $583$ of which $302$ include several components with different types (e.g. a claim followed by premise). 
Therefore, the identification of argument components requires the separation of argumentative from non-argumentative text units and the recognition of component boundaries at the token level.
The proportion of paragraphs with unlinked argument components (e.g. unsupported claims without incoming relations) is $421$ (23\%). 
Thus, methods that link all argument components in a paragraph are only of limited use for identifying the argumentation structures in our corpus.

In total, the corpus includes 1,130 arguments, i.e. claims supported by at least one premise. 
Only 140 of them have an attack relation. 
Thus, the proportion of arguments with attack relations is considerably lower than in the microtext corpus from \namecite{Peldszus2015}.
Most of the arguments are convergent, i.e. the depth of the argument is one. 
The number of arguments with serial structure is $236$ (20.9\%).

\section{Approach}
\label{approach}

\begin{figure}[h]
\centering
\includegraphics[width=1.0\linewidth]{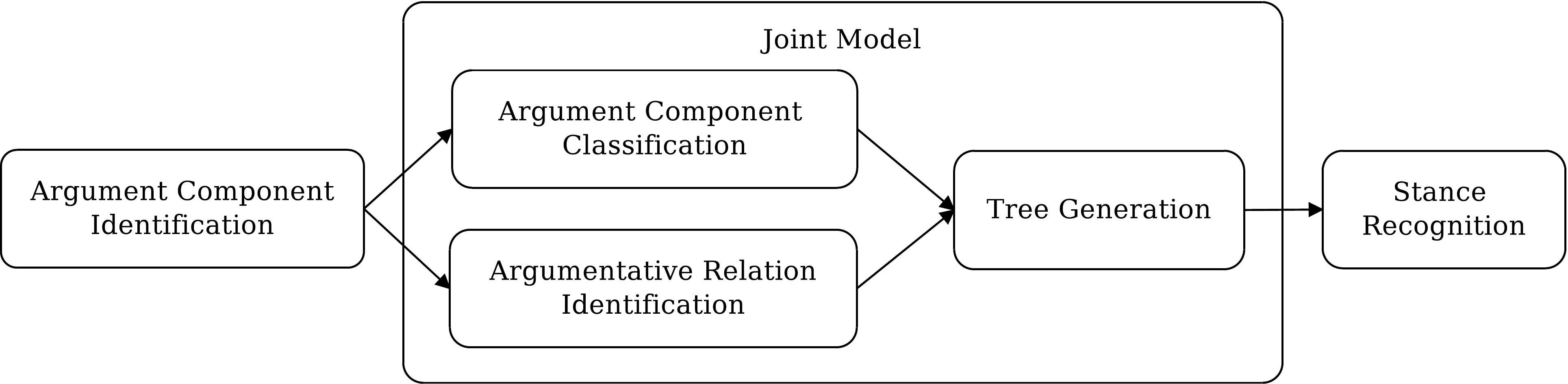}
\caption{Architecture of the argumentation structure parser}
\label{fig:ArchitectureDiagram}
\end{figure}

Our approach for parsing argumentation structures consists of five consecutive subtasks depicted in Figure \ref{fig:ArchitectureDiagram}. 
The \emph{identification model} separates argumentative from non-argumentative text units and recognizes the boundaries of argument components.  
The next three models constitute a joint model for recognizing the argumentation structure. 
We train two base classifiers. The \emph{argument component classification} model labels each argument component as major claim, claim or premise while the \emph{argumentative relation identification} model recognizes if two argument components are argumentatively linked or not. 
The \emph{tree generation} model globally optimizes the results of the two base classifiers for finding a tree (or several ones) in each paragraph. 
Finally, the \emph{stance recognition} model differentiates between support and attack relations.

For preprocessing, we use several models from the \emph{DKPro Framework} \cite{Eckart2014}. 
We identify tokens and sentence boundaries using the LanguageTool segmenter\footnote{www.languagetool.org} and identify paragraphs by checking for line breaks. 
We lemmatize each token using the mate tools lemmatizer \cite{Bohnet2013} and apply the Stanford part-of-speech (POS) tagger \cite{Toutanova2003}, constituent and dependency parsers \cite{Klein2003}, and sentiment analyzer \cite{Socher2013}. 
We use a discourse parser from \namecite{Lin2014} for recognizing PDTB-style discourse relations. 
We employ the \emph{DKPro TC} text classification framework \cite{Daxenberger2014} for feature extraction and experimentation. 

In the following sections, we describe each model in detail. 
For finding the best-performing models, we conduct \emph{model selection} on our training data using 5-fold cross-validation. 
Then, we conduct \emph{model assessment} on our test data. 
We determine the evaluation scores of each cross-validation experiment by accumulating the confusion matrices of each fold into one confusion matrix,  which has been shown to be the less biased method for evaluating cross-validation experiments \cite{Forman2010}. 
We employ macro-averaging as described by \namecite{Sokolova2009} and report macro precision (P), macro recall (R) and macro F1 scores (F1).
We use McNemar test \cite{McNemar1947} with $p=.05$ for significance testing. 
Compared to other tests, it does not make as many assumptions about the distribution in the data \cite{Japkowicz2014}. 
Furthermore, this test compares the outcomes of two classifiers to the gold standard and does not require several trials. 
Thus, it allows for assessing the differences of the models in both of our evaluation scenarios (model selection and model assessment). 

The remainder of this section is structured as follows: In the following section, we introduce the baselines and the upper bound for each task.
In Section \ref{segmentation}, we present the identification model that detects argument components and their boundaries. 
In Section \ref{sec:approachjointmodel}, we propose a new joint model for identifying argumentation structures. 
In Section \ref{stanceClass}, we introduce our stance recognition model. 
In Section \ref{sec:evaluation}, we report the results of the model assessment on our test data and on the microtext corpus from \namecite{Peldszus2015}.
We present the results of the error analysis in Section \ref{sec:errorAnalysis}.
We evaluate the identification model independently and use the gold standard argument components for evaluating the remaining models.

\subsection{Baselines and Upper Bound}
\label{sec:baselines}

For evaluating our models, we use two different types of baselines: 
First, we employ \emph{majority baselines} which label each instance with the majority class. 
Table \ref{tab:classDist} in the appendix shows the class distribution in our training data and test data for each task.

Second, we use \emph{heuristic baselines}, which are motivated by the common structure of persuasive essays \cite{Whitaker2009,Perutz2010}. 
The heuristic baseline of the identification task exploits sentence boundaries. 
It selects all sentences as argument components except the first two and the last sentence of an essay.\footnote{Full stops at the end of a sentence are all classified as non-argumentative.}
The heuristic baseline of the classification task labels the first argument component in each body paragraph as claim and all remaining components in body paragraphs as premise. 
The last argument component in the introduction and the first argument component in the conclusion are classified as major claim and all remaining argument components in the introduction and conclusion are labeled as claim. 
The heuristic baseline for the relation identification classifies an argument component pair as linked if the target is the first component of a body paragraph.
We expect that this baseline will yield good results because $62\%$ of all body paragraphs in our corpus start with a claim.
The heuristic baseline of the stance recognition classifies each argument component in the second last paragraph as attack. 
The motivation for this baseline stems from essay writing guidelines which recommend including opposing arguments in the second last paragraph.

We determine the \emph{human upper bound} for each task by averaging the evaluation scores of all three annotator pairs on our test data.

\subsection{Identifying Argument Components}
\label{segmentation}

We consider the identification of argument components as a sequence labeling task at the token level.
We encode the argument components using an IOB-tagset \cite{Ramshaw1995} and consider an entire essay as a single sequence.
Accordingly, we label the first token of each argument component as ``Arg-B'', the tokens covered by an argument component as ``Arg-I'', and non-argumentative tokens as ``O''. 
As a learner, we use a CRF \cite{Lafferty2001} with averaged perceptron training method \cite{Collins2002}.
Since a CRF considers contextual information, the model is particularly suited for sequence labeling tasks \cite[p.~292]{Goudas2014}. 
For each token, we extract the following features (Table \ref{tab:featuresIdentification}):

\textbf{Structural features} capture the position of the token. 
We expect that these features are effective for filtering non-argumentative text units since the introductions and conclusions of essays include few argumentatively relevant content. 
The punctuation features indicate if the token is a punctuation and if the token  is adjacent to a punctuation. 

\textbf{Syntactic features} consist of the token's POS as well as features extracted from the \emph{lowest common ancestor} (LCA) of the current token $t_i$ and its adjacent tokens in the constituent parse tree.
First, we define ${LCA}_{preceding}(t_i)=\frac{|lcaPath(t_i,t_{i-1})|}{depth}$ where $|lcaPath(u,v)|$ is the length of the path from $u$ to the LCA of $u$ and $v$, and $depth$ the depth of the constituent parse tree. 
Second, we define ${LCA}_{following}(t_i)=\frac{|lcaPath(t_i,t_{i+1})|}{depth}$, which considers the current token $t_i$ and its following token $t_{i+1}$.\footnote{We set $LCA_{preceding} = -1$ if $t_i$ is the first token in its covering sentence and $LCA_{following} = -1$ if $t_i$ is the last token in its covering sentence.} 
Additionally, we add the constituent types of both lowest common ancestors to our feature set. 

\begin{table}[!ht]
\caption{Features used for argument component identification (*indicates genre-dependent features)}\label{tab:featuresIdentification}
\footnotesize{
	\begin{tabularx}{\textwidth}{ | p{0.09\textwidth}  | p{0.27\textwidth} | p{0.54\textwidth} |}

		\cline{1-3}
		\emph{\textbf{Group}} & \emph{\textbf{Feature}} & \emph{\textbf{Description}} \tabularnewline
		\cline{1-3}
		
		\parbox[t]{2mm}{\multirow{8}{*}{\emph{Structural}}} & \parbox[t]{2mm}{\multirow{3}{*}{Token position}} & Token present in introduction or conclusion*; token is   
first or last token in sentence; relative and absolute token position in document, paragraph and sentence\tabularnewline
		\cline{2-3}
		& \parbox[t]{2mm}{\multirow{3}{*}{Punctuation}} & Token precedes or follows any punctuation, full stop, comma and semicolon; token is any punctuation or full stop\tabularnewline
		\cline{2-3}
		& \parbox[t]{2mm}{\multirow{2}{*}{Position of covering sentence}} & Absolute and relative position of the token's covering sentence in the document and paragraph \tabularnewline
		\cline{1-3}
		
		\parbox[t]{2mm}{\multirow{5}{*}{\emph{Syntactic}}} & Part-of-speech & The token's part-of-speech \tabularnewline
		\cline{2-3}
		& Lowest common ancestor (LCA) & Normalized length of the path to the LCA with the following and preceding token in the parse tree \tabularnewline
		\cline{2-3}
		& \parbox[t]{2mm}{\multirow{2}{*}{LCA types}} & The two constituent types of the LCA of the current token and its preceding and following token \tabularnewline
		\cline{1-3}
		
		\parbox[t]{2mm}{\multirow{2}{*}{\emph{LexSyn}}} & \parbox[t]{2mm}{\multirow{2}{*}{Lexico-syntactic}} & Combination of lexical and syntactic features as described by \namecite{Soricut2003} \tabularnewline
		\cline{1-3}
		
		\parbox[t]{2mm}{\multirow{2}{*}{\emph{Prob}}} & \parbox[t]{2mm}{\multirow{2}{*}{Probability}}  & Conditional probability of the current token being the beginning of a component given its preceding tokens \tabularnewline
		\cline{1-3}
	\end{tabularx}
	}
\end{table} 

\textbf{Lexico-syntactic features} have been shown to be effective for segmenting elementary discourse units \cite{Hernault2010}.
We adopt the features introduced by \namecite{Soricut2003}. 
We use lexical head projection rules \cite{Collins2003} implemented in the Stanford tool suite to lexicalize the constituent parse tree. 
For each token $t$, we extract its uppermost node $n$ in the parse tree with the lexical head $t$ and define a lexico-syntactic feature as the combination of $t$ and the constituent type of $n$. 
We also consider the child node of $n$ in the path to $t$ and its right sibling, and combine their lexical heads and constituent types as described by \namecite{Soricut2003}.

The \textbf{probability feature} is the conditional probability of the current token $t_i$ being the beginning of an argument component (``Arg-B'') given its preceding tokens. We maximize the probability for preceding tokens of a length up to $n=3$:
\begin{equation*}
\argmax_{n\in\{1,2,3\}}  P(t_i=\text{\emph{``Arg-B''}}|t_{i-n},...,t_{i-1})
\end{equation*}
To estimate these probabilities, we divide the number of times the preceding tokens $t_{i-n},...,t_{i-1}$ with $1\leq n \leq3$ precede a token $t_i$ labeled as ``Arg-B'' by the total number of occurrences of the preceding tokens in our training data.

\subsubsection{Results of Argument Component Identification}

The results of model selection show that using all features performs best. 
Table \ref{tab:modelSelectionSeg} in the appendix shows the detailed results of the feature analysis. 
Table~\ref{tab:segTestResults} shows the results of the model assessment on the test data. 
The heuristic baseline achieves a macro F1 score of $.642$ and outperforms the majority baseline by $.383$. 
It achieves an F1 score of $.677$ for non-argumentative tokens (``O'') and $.867$ for argumentative tokens (``Arg-I''). 
Thus, the heuristic baseline effectively separates argumentative from non-argumentative text units. 
However, it achieves a low F1 score of $.364$ for identifying the beginning of argument components (``Arg-B''). 
Since it does not split sentences, it recognizes 145 fewer argument components compared to the number of gold standard components in the test data. 

\begin{table}[!ht]

\caption{Model assessment of argument component identification ($\dagger$ = significant improvement over baseline heuristic)}\label{tab:segTestResults}
\footnotesize{
	\begin{tabularx}{\textwidth}{ p{0.25\textwidth} | p{0.07\textwidth} p{0.07\textwidth}  p{0.07\textwidth} | p{0.1\textwidth} p{0.1\textwidth} p{0.1\textwidth} }
	 &  \centering{\emph{F1}} & \centering{\emph{P}} & \centering{\emph{R}} & \centering{\emph{F1 Arg-B}} & \centering{\emph{F1 Arg-I}} & \centering{\emph{F1 O}} 
\tabularnewline
\hline
\emph{Human upper bound} &  \centering{.886} & \centering{.887} & \centering{.885} & \centering{.821} & \centering{.941} & \centering{.892} \tabularnewline
\hline
\emph{Baseline majority} &  \centering{.259} & \centering{.212} & \centering{.333} & \centering{0} & \centering{.778} & \centering{0} \tabularnewline
\emph{Baseline heuristic}  & \centering{.642} & \centering{.664} & \centering{.621} & \centering{.364} & \centering{.867} & \centering{.677} \tabularnewline
\hline
\emph{CRF all features} $\dagger$ & \centering{\textbf{.867}} & \centering{\textbf{.873}} & \centering{\textbf{.861}} & \centering{\textbf{.809}} & \centering{\textbf{.934}} & \centering{\textbf{.857}} \tabularnewline
	\end{tabularx}	
	}
\end{table} 

The CRF model with all features significantly outperforms the heuristic baseline (Table \ref{tab:segTestResults}).
It achieves a macro F1 score of $.867$. 
Compared to the heuristic baseline, it performs considerably better in identifying the beginning of argument components.  
It also performs better for separating argumentative from non-argumentative text units. 
In addition, the number of identified argument components differs only slightly from the number of gold standard components in our test data. 
It identifies $1{,}272$ argument components, whereas the number of gold standard components in our test data amounts to $1{,}266$. 
The human upper bound yields a macro F1 score of $.886$ for identifying argument components. 
The macro F1 score of our model is only $.019$ less. 
Therefore, our model achieves $97.9\%$ of human performance.

\subsubsection{Error Analysis}

\noindent
For identifying the most frequent errors of our model, we manually investigated the predicted argument components. 
The most frequent errors are false positives of ``Arg-I''. 
The model classifies 1,548 out of 9,403 non-argumentative tokens (``O'') as argumentative (``Arg-I''). 
The reason for these errors is threefold: 
First, the model frequently labels non-argumentative sentences in the conclusion of an essay as argumentative.
These sentences are, for instance, non-argumentative recommendations for future actions or summarizations of the essay topic. 
Second, the model does not correctly recognize non-argumentative sentences in body paragraphs. 
It wrongly identifies argument components in $13$ out of the $15$ non-argumentative body paragraph sentences in our test data. 
The reason for these errors may be attributed to the high class imbalance in our training data. 
Third, the model tends to annotate lengthy non-argumentative preceding tokens as argumentative. 
For instance, it labels subordinate clauses preceding the actual argument component as argumentative in sentences similar to ``\emph{\ul{In addition to the reasons mentioned above}, [actual `Arg-B'] ...}'' (underlined text units represent the annotations of our model).

The second most frequent cause of errors are misclassified beginnings of argument components. 
The model classifies 137 of the 1,266 beginning tokens as ``Arg-I''. 
The model, for instance, fails to identify the correct beginning in sentences like ``\emph{Hence, \ul{from this case we are capable of stating that [actual `Arg-B'] ... }}'' or ``\emph{\ul{Apart from the reason I mentioned above, another equally important aspect is that [actual `Arg-B'] ...}}''. 
These examples also explain the false negatives of non-argumentative tokens which are wrongly classified as ``Arg-B''.

\subsection{Recognizing Argumentation Structures}
\label{sec:approachjointmodel}

The identification of argumentation structures involves the classification of argument component types and the identification of argumentative relations.
Both argumentative types and argumentative relations share mutual information \cite[p.~54]{Stab2014b}. 
For instance, if an argument component is classified as claim, it is less likely to exhibit outgoing relations and more likely to have incoming relations. 
On the other hand, an argument component with an outgoing relation and few incoming relations is more likely to be a premise. 
Therefore, we propose a joint model which combines both types of information for finding the optimal structure.
We train two local base classifiers. One classifier recognizes the type of argument components, and another identifies argumentative relations between argument components. 
For both models, we use an SVM \cite{Cortes1995} with a polynomial kernel implemented in the Weka machine learning framework \cite{Hall2009}.
The motivation for selecting this learner stems from the results of our previous work, in which we found that SVMs outperform several other learners in both tasks \cite[p.~51]{Stab2014b}. 
We globally optimize the outcomes of both classifiers in order to find the optimal argumentation structure using integer linear programming. 

\subsubsection{Classifying Argument Components}
\label{compClass}

\noindent
We consider the classification of argument component types as multiclass classification and label each argument component as ``major claim'', ``claim'' or ``premise''. We experiment with the following feature groups: 

\textbf{Lexical features} consist of binary lemmatized unigrams and the 2k most frequent dependency word pairs. We extract the unigrams from the component and its preceding tokens to ensure that discourse markers are included in the features.

\textbf{Structural features} capture the position of the component in the document and token statistics (Table \ref{tab:featuresClassification}). 
Since major claims occur frequently in introductions or conclusions, we expect that these features are valuable for differentiating component types. 

\textbf{Indicator features} are based on four categories of lexical indicators that we manually extracted from $30$ additional essays.
\emph{Forward indicators} such as ``therefore'', ``thus'', or ``consequently'' signal that the component following the indicator is a result of preceding argument components.
\emph{Backward indicators} indicate that the component following the indicator supports a preceding component. Examples of this category are ``in addition'', ``because'', or ``additionally''. 
\emph{Thesis indicators} such as ``in my opinion'' or ``I believe that'' indicate major claims. 
\emph{Rebuttal indicators} signal attacking premises or contra arguments. Examples are ``although'', ``admittedly'', or ``but''.
The complete lists of all four categories are provided in Table \ref{tab:listOfIndicators} in the appendix.
We define for each category a binary feature that indicates if an indicator of a category is present in the component or its preceding tokens. 
An additional binary feature indicates if first-person indicators are present in the argument component or its preceding tokens (Table~\ref{tab:featuresClassification}).
We assume that first-person indicators are informative for identifying major claims.

\textbf{Contextual features} capture the context of an argument component. 
We define eight binary features set to true if a forward, backward, rebuttal or thesis indicator precedes or follows the current component in its covering paragraph. 
Additionally, we count the number of noun and verb phrases of the argument component that are also present in the introduction or conclusion of the essay. 
These features are motivated by the observation that claims frequently restate entities or phrases of the essay topic.
Furthermore, we add four binary features indicating if the current component shares a noun or verb phrase with the introduction or conclusion. 

\textbf{Syntactic features} consist of the POS distribution of the argument component, the number of subclauses in the covering sentence, the depth of the constituent parse tree of the covering sentence, the tense of the main verb of the component, and a binary feature that indicates whether a modal verb is present in the component.

The \textbf{probability features} are the conditional probabilities of the current component being assigned the type $t \in \{MajorClaim, Claim, Premise\}$ given the sequence of tokens $p$ directly preceding the component. 
To estimate $P(t|p)$, we divide the number of times the preceding tokens $p$ appear before a component tagged as $t$ by the total number of occurrences of $p$ in our training data.

\textbf{Discourse features} are based on the output of the PDTB-style discourse parser from \namecite{Lin2014}. 
Each binary feature is a triple combining the following information: (1) the type of the relation that overlaps with the current argument component, (2) whether the current argument component overlaps with the first or second elementary discourse unit of a relation, and (3) if the discourse relation is implicit or explicit. 
For instance, the feature ``\emph{Contrast\_imp\_Arg1}'' indicates that the current component overlaps with the first discourse unit of an implicit contrast relation. 
The use of these features is motivated by the findings of \namecite{Cabrio2013}. 
By analyzing several example arguments, they hypothesized that general discourse relations could be informative for identifying argument components.

\begin{table}[!ht]
\caption{Features of the argument component classification model (*indicates genre-dependent features)}\label{tab:featuresClassification}
\footnotesize{
	\begin{tabularx}{\textwidth}{ | p{0.09\textwidth}  | p{0.27\textwidth} | p{0.54\textwidth} |}

		\cline{1-3}
		\emph{\textbf{Group}} & \emph{\textbf{Feature}} & \emph{\textbf{Description}} \tabularnewline
		\cline{1-3}
		
		\parbox[t]{2mm}{\multirow{3}{*}{\emph{Lexical}}} & \parbox[t]{2mm}{\multirow{2}{*}{Unigrams}} & Binary and lemmatized unigrams of the component and its preceding tokens\tabularnewline
		\cline{2-3}
		& Dependency tuples & Lemmatized dependency tuples (2k most frequent) \tabularnewline
		\cline{1-3}
		
		\parbox[t]{2mm}{\multirow{8}{*}{\emph{Structural}}} & \parbox[t]{2mm}{\multirow{4}{*}{Token statistics}} & Number of tokens of component, covering paragraph and covering sentence; number of tokens preceding and following the component in its sentence; ratio of component and sentence tokens \tabularnewline
		\cline{2-3}
		& \parbox[t]{2mm}{\multirow{4}{*}{Component position}} & Component is first or last in paragraph; component present in introduction or conclusion*; Relative position in paragraph; number of preceding and following components in paragraph \tabularnewline
		\cline{1-3}
		
		\parbox[t]{2mm}{\multirow{4}{*}{\emph{Indicators}}} & \parbox[t]{2mm}{\multirow{2}{*}{Type indicators}} & Forward, backward, thesis or rebuttal indicators present in the component or its preceding tokens \tabularnewline
		\cline{2-3}
		& \parbox[t]{2mm}{\multirow{2}{*}{First-person indicators}} & ``I'', ``me'', ``my'', ``mine'', or ``myself'' present in component or its preceding tokens\tabularnewline
		\cline{1-3}

		\parbox[t]{2mm}{\multirow{4}{*}{\emph{Contextual}}} & \parbox[t]{2mm}{\multirow{2}{*}{Type indicators in context}} & Forward, backward, thesis or rebuttal indicators  preceding or following the component in its paragraph\tabularnewline
		\cline{2-3}
		& \parbox[t]{2mm}{\multirow{2}{*}{Shared phrases*}} & Shared noun phrases or verb phrases with the introduction or conclusion (number and binary) \tabularnewline
		\cline{1-3}
		
		\parbox[t]{2mm}{\multirow{5}{*}{\emph{Syntactic}}} & Subclauses & Number of subclauses in the covering sentence \tabularnewline
		\cline{2-3}
		& Depth of parse tree & Depth of the parse tree of the covering sentence \tabularnewline
		\cline{2-3}
		& Tense of main verb & Tense of the main verb of the component \tabularnewline
		\cline{2-3}
		& Modal verbs & Modal verbs present in the component \tabularnewline
		\cline{2-3}
		& POS distribution & POS distribution of the component \tabularnewline
		\cline{1-3}
		
		\parbox[t]{2mm}{\multirow{2}{*}{\emph{Probability}}} &  \parbox[t]{2mm}{\multirow{2}{*}{Type probability}} & Conditional probability of the component being a major claim, claim or premise given its preceding tokens \tabularnewline
		\cline{1-3}
		
		\parbox[t]{2mm}{\multirow{2}{*}{\emph{Discourse}}} & \parbox[t]{2mm}{\multirow{2}{*}{Discourse Triples}} & PDTB-discourse relations overlapping with the current component\tabularnewline
		\cline{1-3}
		
		\parbox[t]{2mm}{\multirow{2}{*}{\emph{Embedding}}} & \parbox[t]{2mm}{\multirow{2}{*}{Combined word embeddings}} & Sum of the word vectors of each word of the component and its preceding tokens\tabularnewline
		\cline{1-3}
	\end{tabularx}
	}
\end{table} 

\textbf{Embedding features} are based on word embeddings trained on a part of the Google news data set \cite{Mikolov2013}. 
We sum the vectors of each word of an argument component and its preceding tokens and add it to our feature set.
In contrast to common bag-of-words representations, embedding features have a continuous feature space that helped to achieve better results in several NLP tasks \cite{Socher2013}.

By experimenting with individual features and several feature combinations, we found that a combination of all features yields the best results. The results of the model selection can be found in Table \ref{tab:modelSelectionComps} in the appendix.

\subsubsection{Identifying Argumentative Relations} 
\label{relClass}

The relation identification model classifies ordered pairs of argument components as ``linked'' or ``not-linked''. 
In this analysis step, we consider both argumentative support and attack relations as ``linked''. 
For each paragraph with argument components $c_1,...,c_n$, we consider $p = (c_i,c_j)$ with $i \neq j$ and $1\leq i,j\leq n$ as an argument component pair. 
An argument component pair is ``linked'' if our corpus contains an argumentative relation with $c_i$ as source component and $c_j$ as target component. The class distribution is skewed towards ``not-linked'' pairs (Table \ref{tab:classDist}).
We experiment with the following features:

\textbf{Lexical features} are binary lemmatized unigrams of the source and target component and their preceding tokens. We limit the number of unigrams for both source and target component to the $500$ most frequent words in our training data.

\textbf{Syntactic features} include binary POS features of the source and target component and the $500$ most frequent production rules extracted from the parse tree of the source and target component as described in our previous work \cite{Stab2014b}. 

\textbf{Structural features} consist of the number of tokens in the source and target component, statistics on the components of the covering paragraph of the current pair, and position features (Table \ref{tab:featuresRelations}). 

\textbf{Indicator features} are based on the forward, backward, thesis and rebuttal indicators introduced in Section \ref{compClass}. 
We extract binary features from the source and target component and the context of the current pair (Table \ref{tab:featuresRelations}).
We assume that these features are helpful for modeling the direction of argumentative relations and the context of the current component pair. 

\textbf{Discourse features} are extracted from the source and target component of each component pair as described in Section {\ref{compClass}. 
Although PDTB-style discourse relations are limited to adjacent relations, we expect that the types of general discourse relations can be helpful for identifying argumentative relations. 
We also experimented with features capturing PDTB relations between the target and source component. 
However, those were not effective for capturing argumentative relations.

\textbf{PMI features} are based on the assumption that particular words indicate incoming or outgoing relations. 
For instance, tokens like ``therefore'', ``thus'', or ``hence'' can signal incoming relations, whereas tokens such as ``because'', ``since'', or ``furthermore'' may indicate outgoing relations. 
To capture this information, we use \emph{pointwise mutual information} (PMI) which has been successfully used for measuring word associations \cite{Turney2002,Church1990}. 
However, instead of determining the PMI of two words, we estimate the PMI between a lemmatized token $t$ and the direction of a relation $d=\{incoming, outgoing\}$ as $PMI(t,d)=log\frac{p(t,d)}{p(t) \,p(d)}$. 
Here, $p(t,d)$ is the probability that token $t$ occurs in an argument component with either incoming or outgoing relations. 
The ratio between $p(t,d)$ and $p(t)\,p(d)$ indicates the dependence between a token and the direction of a relation.
We estimate $PMI(t,d)$ for each token in our training data.
We extract the ratio of tokens positively and negatively associated with incoming or outgoing relations for both source and target component.
Additionally, we extract four binary features which indicate if any token of the components has a positive or negative association with either incoming or outgoing relations. 

\textbf{Shared noun features (shNo)} indicate if the source and target component share a noun.
We also add the number of shared nouns to our feature set. 
These features are motivated by the fact that premises and claims in classical syllogisms share the same subjects \cite[p.~199]{Govier2010}.

\begin{table}[!ht]
\caption{Features used for argumentative relation identification (*indicates genre-dependent features)}\label{tab:featuresRelations}
\footnotesize{
	\begin{tabularx}{\textwidth}{ | p{0.09\textwidth}  | p{0.28\textwidth} | p{0.53\textwidth} |}

		\cline{1-3}
		\emph{\textbf{Group}} & \emph{\textbf{Feature}} & \emph{\textbf{Description}} \tabularnewline
		\cline{1-3}
		
		\parbox[t]{2mm}{\multirow{3}{*}{\emph{Lexical}}} & \parbox[t]{2mm}{\multirow{3}{*}{Unigrams}} & Binary lemmatized unigrams of the source and target components including preceding tokens (500 most frequent)\tabularnewline
		
		\cline{1-3}
		
		\parbox[t]{2mm}{\multirow{3}{*}{\emph{Syntactic}}} & Part-of-speech & Binary POS features of source and target components\tabularnewline
		\cline{2-3}
		& \parbox[t]{2mm}{\multirow{2}{*}{Production rules}} & Production rules extracted from the constituent parse tree (500 most frequent) \tabularnewline
		\cline{1-3}

		\parbox[t]{2mm}{\multirow{6}{*}{\emph{Structural}}} & Token statistics & Number of tokens of source and target \tabularnewline
		\cline{2-3}
		& \parbox[t]{2mm}{\multirow{2}{*}{Component statistics}} & Number of components between source and target; number of components in covering paragraph \tabularnewline
		\cline{2-3}
		& \parbox[t]{2mm}{\multirow{3}{*}{Position features}} & Source and target present in same sentence; target present before source; source and target are first or last component in paragraph; pair present in introduction or conclusion* \tabularnewline
		\cline{1-3}
		
		\parbox[t]{2mm}{\multirow{4}{*}{\emph{Indicator}}} & Indicator source/target & Indicator type present in source or target \tabularnewline
		\cline{2-3}
		& Indicators between & Indicator type present between source or target \tabularnewline
		\cline{2-3}
		& \parbox[t]{2mm}{\multirow{2}{*}{Indicators context}} & Indicator type follows or precedes source or target in the covering paragraph of the pair\tabularnewline
		\cline{1-3}

		\parbox[t]{2mm}{\multirow{1}{*}{\emph{Discourse}}} & Discourse Triples  & Binary discourse triples of source and target \tabularnewline
		\cline{1-3}

		\parbox[t]{2mm}{\multirow{4}{*}{\emph{PMI}}} &  \parbox[t]{2mm}{\multirow{4}{*}{Pointwise mutual information}} & Ratio of tokens positively or negatively associated with incoming or outgoing relations; Presence of words negatively or positively associated with incoming or outgoing relations\tabularnewline
		\cline{1-3}

		\parbox[t]{2mm}{\multirow{2}{*}{\emph{ShNo}}} & \parbox[t]{2mm}{\multirow{2}{*}{Shared nouns}} & Shared nouns between source and target components (number and binary)\tabularnewline
		\cline{1-3}

	\end{tabularx}
	}
\end{table} 

For selecting the best performing model, we conducted feature ablation tests and experimented with individual features. 
The results show that none of the feature groups is informative when used individually. 
We achieved the best performance by removing lexical features from our feature set 
(detailed results of the model selection can be found in Table \ref{tab:modelSelectionRel} in the appendix).

\subsubsection{Jointly Modeling Argumentative Relations and Argument Component Types}
\label{jointModel}

Both base classifiers identify argument component types and argumentative relations locally. 
Consequently, the results may not be globally consistent.
For instance, the relation identification model does not link 37.1\% of all premises in our model selection experiments.
Therefore, we propose a joint model that globally optimizes the outcomes of the two base classifiers. 
We formalize this task as an \emph{integer linear programming} (ILP) problem. 
Given a paragraph including $n$ argument components\footnote{We consider only claims and premises in our joint model since argumentative relations between claims and major claims are modeled with a level approach (cf. Section \ref{annotationScheme}).}, we define the following objective function
\begin{equation}
\label{objectiveFunction}
\argmax_x \sum_{i=1}^{n} \sum_{j=1}^{n} w_{ij} x_{ij}
\end{equation}
\noindent
with variables $x_{ij} \in \{0,1\}$ indicating an argumentative relation from argument component $i$ to argument component $j$.\footnote{We use the lpsolve framework (http://lpsolve.sourceforge.net) and set each variable in the objective function to ``binary mode'' for ensuring the upper bound of 1.} 
Each coefficient $w_{ij} \in \mathbb{R}$ is a weight of a relation. 
It is determined by incorporating the outcomes of the two base classifiers. 
For ensuring that the resulting structure is a tree, we define the following constraints: 
\begin{equation}
  \label{const_outRelations}
  \forall i : \sum_{j=1}^{n} x_{ij} \leq 1 
\end{equation}
\vspace{-0.8cm}
\begin{equation}
  \label{const_oneOrLessRoots}
\sum_{i=1}^{n} \sum_{j=1}^{n} x_{ij} \leq n-1
\end{equation}
\vspace{-0.7cm}
\begin{equation}
  \label{const_selfRelations}
  \forall i : x_{ii} = 0 
\end{equation}

\noindent
Equation \ref{const_outRelations} prevents an argument component $i$ from having more than one outgoing relation. 
Equation \ref{const_oneOrLessRoots} ensures that a paragraph includes at least one root node, i.e. a node without outgoing relation. 
Equation \ref{const_selfRelations} prevents an argumentative relation from having the same source and target component. 

For preventing cycles, we adopt the approach described by \namecite[p. 92]{Kuebler2009}.
We add the auxiliary variables $b_{ij} \in \{0,1\}$ to our objective function (\ref{objectiveFunction}) where $b_{ij} = 1$ if there is a directed path from argument component $i$ to argument component $j$. 
The following constraints tie the auxiliary variables $b_{ij}$ to the variables $x_{ij}$:
\begin{equation}
  \label{const_cycle1}
  \forall i \forall j : x_{ij} - b_{ij} \leq 1
\end{equation}
\vspace{-0.9cm}
\begin{equation}
  \label{const_cycle2}
  \forall i \forall j \forall k : b_{ik} - b_{ij} - b_{jk} \leq -1
\end{equation}
\vspace{-0.9cm}
\begin{equation}
  \label{const_cycle3}
  \forall i : b_{ii} = 0
\end{equation}
\noindent
The first constraint ensures that there is a path from $i$ to $j$ represented in variable $b_{ij}$ if there is a direct relation between the argument components $i$ and $j$. 
The second constraint covers all paths of length greater than $1$ in a transitive way. 
It states that if there is a path from argument component $i$ to argument component $j$ ($b_{ij} = 1$) and another path from argument component $j$ to argument component $k$ ($b_{jk} = 1$) then there is also a path from argument component $i$ to argument component $k$. 
Thus, it iteratively covers paths of length $l+1$ by having covered paths of length $l$. 
The third constraint prevents cycles by preventing all directed paths starting and ending with the same argument component.

Having defined the ILP model, we consolidate the results of the two base classifiers. 
We consider this task by determining the \emph{weight matrix} $W \in \mathbb{R}^{n \times n}$ that includes the coefficients $w_{ij} \in W$ of our objective function. 
The weight matrix $W$ can be considered an adjacency matrix. 
The greater a weight of a particular relation is, the higher the likelihood that the relation appears in the optimal structure found by the ILP-solver.

First, we incorporate the results of the relation identification model. 
Its result can be considered as an adjacency matrix $R \in {\{0,1\}^{n \times n}}$. 
For each pair of argument components $(i,j)$ with $1 \leq i,j \leq n$, each $r_{ij} \in R$ is $1$ if the relation identification model predicts an argumentative relation from argument component $i$ (source) to argument component $j$ (target), or $0$ if the model does not predict an argumentative relation. 

Second, we derive a \emph{claim score} (cs) for each argument component $i$ from the predicted relations in $R$:
\begin{equation}
\label{claimProb}
	cs_i = \frac{relin_i - relout_i + n-1}{rel + n - 1}
\end{equation}
\noindent
Here, $relin_i = \sum_{k=1}^{n} r_{ki} [i \neq k]$ is the number of predicted incoming relations of argument component $i$, $relout_i = \sum_{l=1}^{n} r_{il} [i \neq l]$ is the number of predicted outgoing relations of argument component $i$ and $rel=\sum_{k=1}^{n}\sum_{l=1}^{n}r_{kl}[k \neq l]$ is the total number of relations predicted in the current paragraph. 
The claim score $cs_i$ is greater for argument components with many incoming relations and few outgoing relations. 
It becomes smaller for argument components with fewer incoming relations and more outgoing relations. 
By normalizing the score with the total number of predicted relations and argument components, it also accounts for contextual information in the current paragraph and prevents overly optimistic scores. 
For example, if all predicted relations point to argument component $i$ which has no outgoing relations, $cs_i$ is exactly $1$. 
On the other hand, if there is an argument component $j$ with no incoming and one outgoing relation in a paragraph with $4$ argument components and $3$ predicted relations in $R$, $cs_j$ is $\frac{1}{3}$. 
Since it is more likely that a relation links an argument component which has a lower claim score to an argument component with a higher claim score, we determine the weight for each argumentative relation as: 
\begin{equation}
cr_{ij} = cs_j - cs_i
\end{equation}
\noindent
By adding the claim score $cs_{j}$ of the target component $j$, we assign a higher weight to relations pointing to argument components which are likely to be a claim. 
By subtracting the claim score $cs_i$ of the source component $i$, we assign smaller weights to relations outgoing argument components with larger claim score. 

Third, we incorporate the argument component types predicted by the classification model. 
We assign a higher score to the weight $w_{ij}$ if the target component $j$ is predicted as claim since it is more likely that argumentative relations point to claims. 
Accordingly, we set $c_{ij} = 1$ if argument component $j$ is labeled as claim and $c_{ij} = 0$ if argument component $j$ is labeled as premise.

Finally, we combine all three scores to estimate the weights of the objective function:
\begin{equation}
\label{eq:weightEstimation}
w_{ij} = \phi_r r_{ij} + \phi_{cr} cr_{ij} + \phi_{c} c_{ij}
\end{equation}
Each $\phi$ represents a hyperparameter of the ILP model. 
In our model selection experiments, we found that 
$\phi_r = \frac{1}{2}$ and $\phi_{cr} = \phi_{c} = \frac{1}{4}$ yields the best performance. 
More detailed results of the model selection are provided in Table \ref{tab:modelSelectionILP} in the appendix.

After applying the ILP model, we adapt the argumentative relations and argument  types according to the results of the ILP-solver.
We revise each relation according to the determined $x_{ij}$ scores, set the type of all components without outgoing relation to ``claim'', and set the type of all remaining components to ``premise''.

\subsection{Classifying Support and Attack Relations}
\label{stanceClass}

The stance recognition model differentiates between argumentative support and attack relations. 
We model this task as binary classification and classify each claim and premise as ``support'' or ``attack''. 
The stance of each premise is encoded in the type of its outgoing relation, whereas the stance of each claim is encoded in its stance attribute. 
We use an SVM and the following features (Table \ref{tab:featuresStance})\footnote{For finding the best learner, we compared Na\"{i}ve Bayes \cite{John1995}, Random Forests \cite{Breiman2001}, Multinomial Logistic Regression \cite{leCessie1992}, C4.5 Decision Trees \cite{Quinlan1993} and SVM \cite{Cortes1995} and found that an SVM considerably outperforms all other classifiers.}:

\textbf{Lexical features} are binary lemmatized unigram features of the argument component and its preceding tokens. 

\textbf{Sentiment features} are based on the subjectivity lexicon from \namecite{Wilson2005} and the five sentiment scores produced by the Stanford sentiment analyzer \cite{Socher2013}. 

\textbf{Syntactic features} consist of the POS distribution of the component and production rules \cite{Stab2014b}.

\textbf{Structural features} capture the position of the component in the paragraph and token statistics (Table \ref{tab:featuresStance}).

\textbf{Discourse features} are discourse triples as described in Section \ref{compClass}.
We expect that these features will be helpful for identifying attacking components since the PDTB includes contrast and concession relations.

\textbf{Embedding features} are the embedding features described in Section \ref{compClass}.

\begin{table}[!ht]
\caption{Features used for stance recognition}\label{tab:featuresStance}
\footnotesize{
	\begin{tabularx}{\textwidth}{ | p{0.09\textwidth}  | p{0.27\textwidth} | p{0.54\textwidth} | }

		\cline{1-3}
		\emph{\textbf{Group}} & \emph{\textbf{Feature}} & \emph{\textbf{Description}} \tabularnewline
		\cline{1-3}
		
		\parbox[t]{2mm}{\multirow{2}{*}{\emph{Lexical}}} & \parbox[t]{2mm}{\multirow{2}{*}{Unigrams}} & Binary and lemmatized unigrams of the component and its preceding token\tabularnewline
		\cline{1-3}

		\parbox[t]{2mm}{\multirow{5}{*}{\emph{Sentiment}}} & \parbox[t]{2mm}{\multirow{3}{*}{Subjectivity clues}} & Presence of negative words; number of negative, positive, and neutral words; number of positive words subtracted by the number of negative words \tabularnewline
		\cline{2-3}
		& \parbox[t]{2mm}{\multirow{2}{*}{Sentiment scores}} & Five sentiment scores of covering sentence (Stanford sentiment analyzer) \tabularnewline
		\cline{1-3}
		
		\parbox[t]{2mm}{\multirow{2}{*}{\emph{Syntactic}}} & POS distribution & POS distribution of the component \tabularnewline
		\cline{2-3}
		& Production rules & Production rules extracted from the constituent parse tree \tabularnewline
		\cline{1-3}
		
		\parbox[t]{2mm}{\multirow{6}{*}{\emph{Structural}}} & \parbox[t]{2mm}{\multirow{3}{*}{Token statistics}} & Number of tokens of covering sentence; number of preceding and following tokens in covering sentence; ratio of component and sentence tokens \tabularnewline
		\cline{2-3}
		& \parbox[t]{2mm}{\multirow{2}{*}{Component statistics}} & Number of components in paragraph; number of preceding and following components in paragraph \tabularnewline
		\cline{2-3}
		& Component Position & Relative position of the argument component in paragraph \tabularnewline
		\cline{1-3}

		\parbox[t]{2mm}{\multirow{2}{*}{\emph{Discourse}}} & \parbox[t]{2mm}{\multirow{2}{*}{Discourse Triples}} & PDTB discourse relations overlapping with the current component\tabularnewline
		\cline{1-3}

		\parbox[t]{2mm}{\multirow{2}{*}{\emph{Embedding}}} & \parbox[t]{2mm}{\multirow{2}{*}{Combined word embeddings}} & Sum of the word vectors of each word of the component and its preceding tokens \tabularnewline
		\cline{1-3}

	\end{tabularx}
	}
\end{table}

\subsection{Evaluation}
\label{sec:evaluation}

The upper part of Table \ref{tab:results} shows the F1 scores of the classification, relation identification, and stance recognition tasks using our test data.
The heuristic baselines outperform the majority baselines in all three tasks by a considerable margin. 
They achieve an average macro F1 score of .674, which confirms our assumption that argumentation structures in persuasive essays can be identified with simple heuristic rules (Section \ref{sec:baselines}).

Our base classifiers for component classification and relation identification both improve the macro F1 scores of the heuristic baselines. 
The component classification model achieves a macro F1 score of .794. 
Compared to the heuristic baseline, the model yields slightly worse results for claims and premises but improves the identification of major claims by .132.
However, the difference between the component classification model and the heuristic baseline is not statistically significant. 
On the other hand, the relation identification model significantly improves the result of the heuristic baseline, achieving a macro F1 score of .717.
Additionally, the stance recognition model significantly outperforms the heuristic baseline by .118 macro F1 score. 
It yields an F1 score of .947 for supporting components and .413 for attacking component. 

\begin{table}[!ht]
\footnotesize
\setlength{\tabcolsep}{.5em}
\def\arraystretch{1.2}
\caption[F1 scores of model assessment.]{F1 scores of model assessment. The upper part shows the results on the test data of the persuasive essay corpus, while the lower part shows the results on the microtext corpus from \namecite{Peldszus2015} ($\dagger$ = significant improvement over baseline heuristic; $\ddagger$ = significant improvement over base classifier).}\label{tab:results}
	
	\resizebox{\textwidth}{!}{%
	\begin{tabular}{ l | c c c c c | c c c c | c c c c | c }
	 & 	\multicolumn{5}{c|}{\emph{Components}} & \multicolumn{4}{c|}{\emph{Relations}} & \multicolumn{4}{c|}{\emph{Stance recognition}} & \tabularnewline
	 &  & \centering{F1} & \centering{F1 MC} & \centering{F1 Cl} & \centering{F1 Pr} & & \centering{F1} & \centering{F1 NoLi} & \centering{F1 Li} & &\centering{F1} & \centering{F1 Sup} & \centering{F1 Att} & Avg F1 \tabularnewline
	 \hline
	 \multicolumn{15}{c}{\emph{Model assessment on persuasive essays}} \tabularnewline
	\hline
\emph{Human upper bound} & & .868 & .926 & .754 & .924 & & .854 & .954 & .755 & & .844 & .975 & .703 & .855\tabularnewline
\hline
\emph{Baseline majority} & & .260 & 0 & 0 & .780 & & .455 & .910 & 0 & & .478 & \textbf{.957} & 0 & .398 \tabularnewline

\emph{Baseline heuristic} & & .759 & .759 & .620 & .899 & & .700 & .901 & .499 & & .562 & .776 & .201 & .674 \tabularnewline
\hline

\emph{Base classifier} & & .794 & \textbf{.891} & .611 & .879 & $\dagger$ & .717 & .917 & .508 & $\dagger$ & \textbf{.680} & .947 & \textbf{.413} & .730 \tabularnewline

\emph{ILP joint model} & $\dagger\ddagger$ & \textbf{.826} & \textbf{.891} & \textbf{.682} & \textbf{.903} & $\dagger$ & \textbf{.751} & \textbf{.918} & \textbf{.585} & $\dagger$ & \textbf{.680} & .947 & \textbf{.413} & \textbf{.752}\tabularnewline
	
	\hline
	\multicolumn{15}{c}{\emph{Model assessment on microtexts}} \tabularnewline
	\hline
	
	\emph{Simple} & & .817 & - & - & - &  & .663 & - & .478 & & .671 & - & - & .717\tabularnewline

	\emph{Best EG} & & \textbf{.869} & - & - & - & & .693 & - & .502 & & .710 & - & - & .757 \tabularnewline
	
	\emph{MP+p} & & .831 & - & - & - &  & \textbf{.720} & - & .546 & & .514 & - & - & .688 \tabularnewline
\hline
	\emph{Base classifier} & & .830 & - & .712 & .937 & & .650 & .841 & .446 & & \textbf{.745} & .855 & .628 & .742 \tabularnewline
	
	\emph{ILP joint model} & & .857 & - & .770 & .943 & & .683 & .881 & .486 & & \textbf{.745} & .855 & .628 & \textbf{.762} \tabularnewline
	
	\end{tabular}	
	}
\end{table} 

The ILP joint model significantly outperforms the heuristic baselines for component classification and relation identification. 
Additionally, it significantly outperforms the base classifier for component classification.
However, it does not yield a significant improvement over the base classifier for relation identification despite that the ILP joint model improves the base classifier for relation identification by $.034$ macro F1 score. 
The results show that the identification of claims and linked component pairs benefit most from the joint model. 
Compared to the base classifiers, the ILP joint model improves the F1 score of claims by $.071$ and the F1 score of linked component pairs by $.077$.

The human upper bound yields macro F1 scores of $.868$ for component classification, $.854$ for relation identification, and $.844$ for stance recognition. 
The ILP joint model achieves almost human performance for classifying argument components. 
Its F1 score is only $.042$ lower compared to human upper bound.
Regarding relation identification and stance recognition, the F1 scores of our model are $.103$ and $.164$ less than human performance.
Thus, our model achieves $95.2\%$ human performance for component identification, $87.9\%$ for relation identification, and $80.5\%$ for stance recognition.

In order to verify the effectiveness of our approach, we also evaluated the ILP joint model on the English microtext corpus (cf. Setion \ref{existingCorpora}). 
For ensuring the comparability to previous results, we used the same data splitting and the repeated cross-validation setup described by \namecite{Peldszus2015}. 
Since the microtext corpus does not include major claims, we removed the major claim label from our component classification model for this evaluation task. 
Furthermore, it was necessary to adapt several features of the base classifiers since the microtext corpus does not include non-argumentative text units.
Therefore, we did not consider preceding tokens for lexical, indicator and embedding features and removed the probability feature of the component classification model.
Additionally, we removed all genre-dependent features of both base classifiers. 

The first three rows of the lower part in Table \ref{tab:results} show the results reported by \namecite{Peldszus2015} on the English microtext corpus. 
The \emph{simple} model indicates their local base classifiers, \emph{Best EG} is their best model for component classification, and \emph{MP+p} is their best model for relation identification. 
On average our base classifiers outperform the base classifiers from \namecite{Peldszus2015} by $.025$. 
Only their relation identification model yields a better macro F1 score compared to our base classifier. 
Their Best~EG model outperforms our model with respect to component classification and relation identification but yields a lower score for stance recognition.
Their MP+p model outperforms the relation identification of our model, but  yields lower results for component classification and stance recognition compared to our ILP joint model. 
This difference can be attributed to the additional information about the function and role attribute incorporated in their joint models (cf. Section \ref{sec:sotaStructure}). 
They showed that both have a beneficial effect on the component classification and relation identification in their corpus \cite[Figure 3]{Peldszus2015}. 
However, the role attribute is a unique feature of their corpus and the arguments in their corpus exhibit an unusually high proportion of attack relations (cf. Section \ref{existingCorpora}). 
In particular, $86.6\%$ of their arguments include attack relations, whereas the proportion of arguments with attack relations in our corpus amounts to only $12.4\%$. 
This proportion may even be lower in other text genres because essay writing guidelines encourage students to include opposing arguments in their writing. 
Therefore, we assume that incorporating function and role attributes will not be beneficial using our corpus. 

The evaluation results show that our ILP joint model simultaneously improves the performance of component classification and relation identification on both corpora.

\subsection{Error Analysis}
\label{sec:errorAnalysis}

In order to analyze frequent errors of the ILP joint model, we investigated the predicted argumentation structures in our test data. 
The confusion matrix of the component classification task (Table~\ref{tab:compClassConf}) shows that the highest confusion is between claims and premises. 
The model classifies $74$ actual premises as claims and $82$ claims as premises.
By manually investigating these errors, we found that the model tends to label inner premises in serial structures as claims and wrongly identifies claims in sentences containing two premises.
\begin{table}[!ht]
\caption{Confusion matrix of the ILP joint model of component classification on our test data}\label{tab:compClassConf}
\footnotesize{
	\begin{tabularx}{\textwidth}{ p{0.01\textwidth} p{0.17\textwidth}  | p{0.22\textwidth}  p{0.22\textwidth}  p{0.22\textwidth} }
		& & \multicolumn{3}{ c }{\emph{predictions}} \tabularnewline
		&& \centering{MajorClaim} & \centering{Claim} & \centering{Premise}\tabularnewline
		\hline
		\parbox[t]{2mm}{\multirow{3}{*}{\rotatebox[origin=c]{90}{\emph{actual}}}} & MajorClaim & \centering{\textbf{139}} & \centering{12} & \centering{2}  \tabularnewline
		&Claim & \centering{20} & \centering{\textbf{202}} & \centering{82}  \tabularnewline
		&Premise & \centering{0} & \centering{74} & \centering{\textbf{735}}  \tabularnewline
	\end{tabularx}
	}
\end{table} 
Regarding the relation identification, we observed that the model tends to identify argumentation structures which are more shallow than the structures in our gold standard. 
The model correctly identifies only $34.7\%$ of the $98$ serial arguments in our test data.
This can be attributed to the ``claim-centered'' weight calculation in our objective function. 
In particular, the predicted relations in matrix $R$ are the only information about serial arguments, whereas the other two scores (c and cr) assign higher weights to relations pointing to claims.

In order to determine if the ILP joint model correctly models the relationship between component types and argumentative relations, we artificially improved the predictions of both base classifiers as suggested by \namecite{Peldszus2015}. 
The dashed lines in Figure \ref{fig:simulation} show the performance of the artificially improved base classifiers.
Continuous lines show the resulting performance of the ILP joint model. 
\begin{figure}[!ht]
    \centering{
     \subfigure{\includegraphics[width=.9\textwidth]{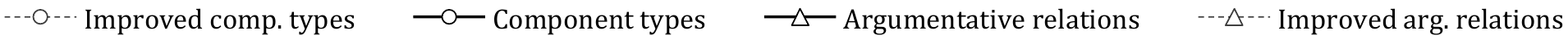}} 
     }
     \addtocounter{subfigure}{-1}
    \centering{
    \subfigure[Improve Types]{\includegraphics[width=0.325\textwidth]{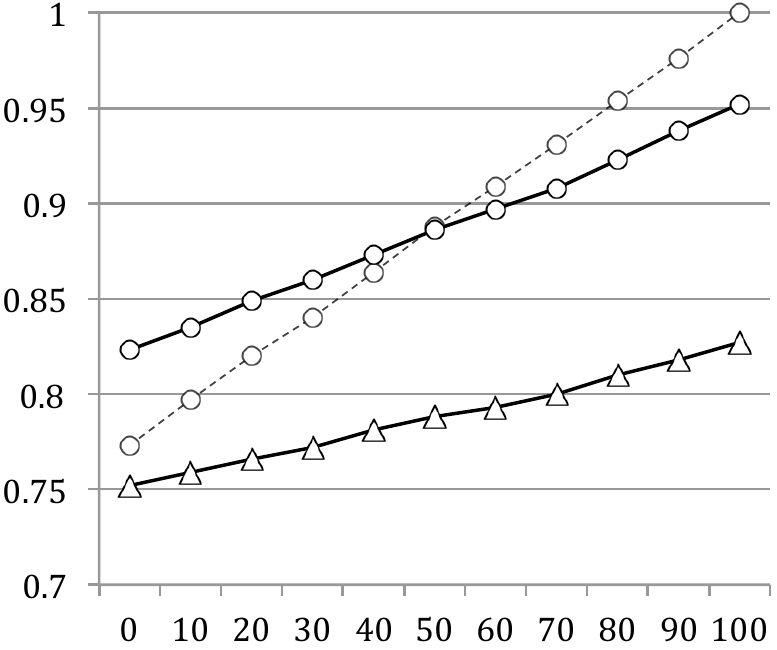}} 
    \subfigure[Improve Relations]{\includegraphics[width=0.325\textwidth]{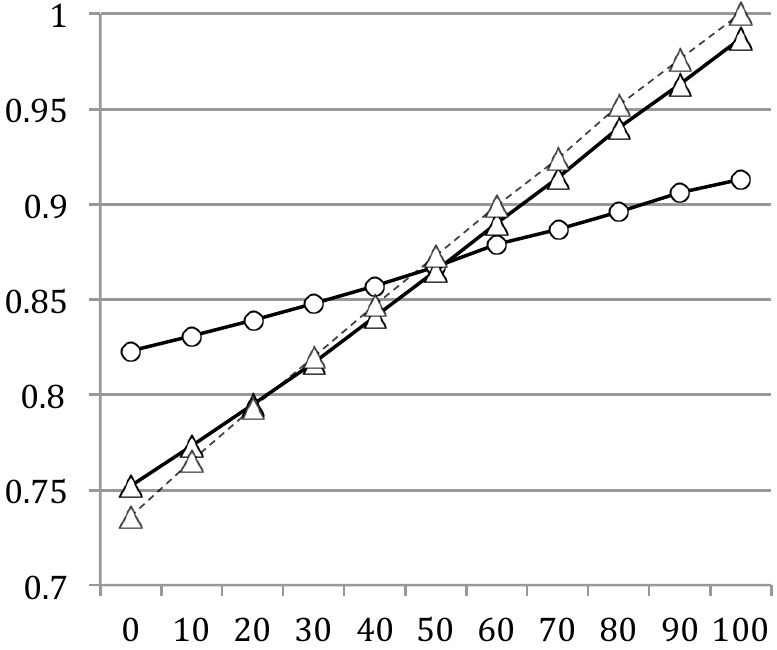}} 
    \subfigure[Improve Both]{\includegraphics[width=0.325\textwidth]{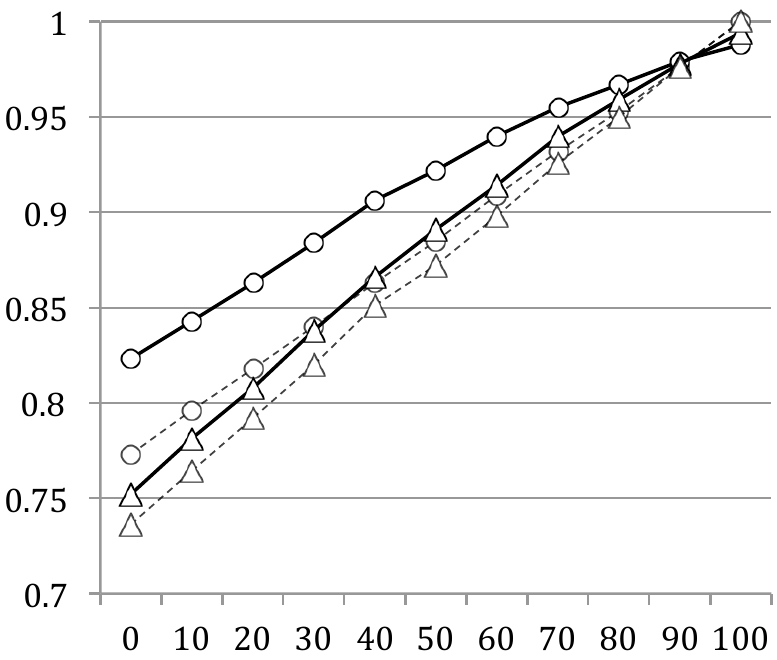}} 
    }
\caption[Influence of improving the base classifiers]{Influence of improving the base classifiers (x-axis shows the proportion of improved predictions and y-axis the macro F1 score).} 
\label{fig:simulation}
\end{figure} 
Figures \ref{fig:simulation}a+b show the effect of improving the component classification and relation identification. 
It shows that correct predictions of one base classifier are not maintained after applying the ILP model if the other base classifier exhibits less accurate predictions. 
In particular, less accurate argumentative relations have a more detrimental effect on the component types (Figure \ref{fig:simulation}a) than less accurate component types do on the outcomes of the relation identification (Figure \ref{fig:simulation}b). 
Thus, it is more reasonable to focus on improving relation identification than component classification in future work. 

Figure \ref{fig:simulation}c depicts the effect of improving both base classifiers, which illustrates that the ILP joint model improves the component types more effectively than argumentative relations. 
Figure \ref{fig:simulation}c also shows that the ILP joint model improves both tasks if the base classifiers are improved. 
Therefore, we conclude that the ILP joint model successfully captures the natural relationship between argument component types and argumentative relations.

\section{Discussion}
\label{discussion}

Our argumentation structure parser includes several consecutive steps. Consequently, potential errors of the upstream models can negatively influence the results of the downstream models. 
For example, errors of the identification model can result in flawed argumentation structures if argumentatively relevant text units are not recognized or non-argumentative text units are identified as relevant. 
However, our identification model yields good accuracy and an $\alpha_U$ of $.958$ for identifying argument components. 
Therefore, it is unlikely that identification errors will significantly influence the outcome of the upstream models when applied to persuasive essays. 
However, as demonstrated by \namecite{Levy2014} and \namecite{Goudas2014}, the identification of argument components is more complex in other text genres than it is in persuasive essays. 
Another potential issue of the pipeline architecture is that wrongly classified major claims will decrease the accuracy of the model due to the fact that those are not integrated in the joint modeling approach.
For this reason, it is worthwhile to experiment in future work with structured machine learning methods which incorporate several tasks in a single model \cite{Moens2014}.

In this work, we have demonstrated that our annotation scheme can be reliably applied to persuasive essays. 
However, persuasive essays exhibit a common structure and so it may be more challenging to apply the annotation scheme to text genres with less explicit argumentation structures such as social media data, product reviews or dialogical debates.
Nevertheless, we believe that our annotation scheme can be successfully applied to other text genres with minor adaptations. 
Although other text genres may not include major claims, previous work has already demonstrated that claims and premises can be reliably annotated in legal cases \cite{MochalesPalau2011}, written dialogs \cite{Biran2011} and even over multiple Wikipedia articles \cite{Aharoni2014}. 
Additionally, it is unknown if our tree assumption generalizes to other text genres. 
Although most previous work considered argumentation structures as trees, other text genres may include divergent arguments and even cyclic argumentation structures. 

Although our approach shows promising results, it is still unknown if the identified argumentation structures can be used to provide adequate feedback about argumentation. 
However, the identified argumentation structures enable various kinds of feedback about argumentation. 
For instance, it facilitates the automatic recommendation of more meaningful and comprehensible argumentation structure. 
Particularly, the extracted structure can be used to prevent multiple reasoning directions in a single argument (e.g. both forward and backward reasoning), which may result in a more comprehensible structure of arguments.
It could be also used to highlight unsupported claims and then prompt the author for reasons supporting or attacking it (e.g. premises related to the claim). 
Additionally, the identified argumentation structure facilitates the recommendation of additional discourse markers in order to make the arguments more coherent or can be used to encourage authors to discuss opposing views. 
Finally, the visualization of the identified argumentation structure could stimulate self reflection and plausibility checking. 
However, finding adequate feedback types and investigating their effect on the argumentation skills of students requires the integration of the models in writing environments and extensive long term user studies in future work.

\section{Conclusion}
\label{conclusion}
In this paper, we presented an end-to-end approach for parsing argumentation structures in persuasive essays. 
Previous approaches suffer from several limitations: 
Existing approaches either focus only on particular subtasks of argumentation structure parsing or rely on manually created rules.
Consequently, previous approaches are only of limited use for parsing argumentation structures in real application scenarios. 
To the best of our knowledge, the presented work is the first approach which covers all required subtasks for identifying the global argumentation structure of documents. 
We showed that jointly modeling argumentation structures simultaneously improves the results of component classification and relation identification. 
Additionally, we introduced a novel annotation scheme and a new corpus of persuasive essays annotated with argumentation structures which represents the largest resource of its kind. 
Both the corpus and the annotation guidelines are freely available in order to ensure reproducibility and for fostering future research in computational argumentation.

\appendix

\appendixsection{Class Distributions}
\label{sec:appendixClassDistribution}

\setcounter{table}{0}
\renewcommand{\thetable}{A\arabic{table}}
Table \ref{tab:classDist} shows the class distributions of the training and test data of the persuasive essay corpus for each analysis step.

\begin{table}[!ht]
\caption{Class distributions in training data and test data}\label{tab:classDist}
\footnotesize{
	\begin{tabularx}{\textwidth}{  p{0.30\textwidth}  | p{0.30\textwidth} | p{0.30\textwidth}  }
		\centering{\emph{\textbf{Class}}} & \centering{\emph{\textbf{Training data}}} & \centering{\emph{\textbf{Test data}}} \tabularnewline
	\hline
	\multicolumn{3}{c}{\emph{Identification}} \tabularnewline
	\hline
	\centering{\emph{Arg-B}} & \centering{4{,}823 (4.1\%)} & \centering{1{,}266 (4.3\%)} \tabularnewline
	\centering{\emph{Arg-I}} & \centering{75{,}053 (63.6\%)} & \centering{18{,}655 (63.6\%)} \tabularnewline
	\centering{\emph{O}} & \centering{38{,}071} (32.3\%) & \centering{9{,}403 (32.1\%)} \tabularnewline
	\hline
	\multicolumn{3}{c}{\emph{Component classification}} \tabularnewline
	\hline
	\centering{\emph{MajorClaim}} & \centering{598 (12.4\%)} & \centering{153 (12.1\%)} \tabularnewline
	\centering{\emph{Claim}} & \centering{1{,}202 (24.9\%)} & \centering{304 (24.0\%)} \tabularnewline
	\centering{\emph{Premise}} & \centering{3{,}023} (62.7\%) & \centering{809 (63.9\%)} \tabularnewline
	\hline
	\multicolumn{3}{c}{\emph{Relation identification}} \tabularnewline
	\hline
	\centering{\emph{Not-Linked}} & \centering{14{,}227 (82.5\%)} & \centering{4{,}113 (83.5\%)} \tabularnewline
	\centering{\emph{Linked}} & \centering{3{,}023 (17.5\%)} & \centering{809 (16.5\%)} \tabularnewline
	\hline
	\multicolumn{3}{c}{\emph{Stance recognition}} \tabularnewline
	\hline
	\centering{\emph{Support}} & \centering{3{,}820 (90.4\%)} & \centering{1{,}021 (91.7\%)} \tabularnewline
	\centering{\emph{Attack}} & \centering{405 (9.6\%)} & \centering{92 (8.3\%)} \tabularnewline
	
	\end{tabularx}
	}
\end{table}

\appendixsection{Detailed Results of Model Selections}
\label{app:modelSelection}

\setcounter{table}{0}
\renewcommand{\thetable}{B\arabic{table}}

The following tables show the model selection results for all five tasks using 5-fold cross-validation on our training data. 
Table \ref{tab:modelSelectionSeg} shows the results of using individual feature groups for the argument component identification task. 
Lexico-syntactic features perform best for identifying argument components, and they perform particularly well for recognizing the beginning of argument components (``Arg-B''). 
The second best features are structural features. 
They yield the best F1 score for separating argumentative from non-argumentative text units (``O'').
\begin{table}[!ht]
\caption{Argument component identification ($\dagger$ = significant improvement over baseline heuristic)}\label{tab:modelSelectionSeg}
\footnotesize{
	\begin{tabularx}{\textwidth}{ p{0.24\textwidth} | p{0.07\textwidth} p{0.07\textwidth}  p{0.07\textwidth} | p{0.1\textwidth} p{0.1\textwidth} p{0.1\textwidth} }
	 & \centering{\emph{F1}} & \centering{\emph{P}} & \centering{\emph{R}} & \centering{\emph{F1 Arg-B}} & \centering{\emph{F1 Arg-I}} & \centering{\emph{F1 O}} 
\tabularnewline
\hline
\emph{Baseline majority}  & \centering{.259} & \centering{.212} & \centering{.333} & \centering{0} & \centering{.778} & \centering{0} \tabularnewline
\emph{Baseline heuristic}  & \centering{.628} & \centering{.647} &\centering{.610} & \centering{.350} & \centering{.869} & \centering{.660} \tabularnewline
\hline
\emph{CRF only structural} $\dagger$  & \centering{.748} & \centering{.757} & \centering{.740}  & \centering{.542} & \centering{.906} & \centering{.789} \tabularnewline
\emph{CRF only syntactic} $\dagger$ & \centering{.730} & \centering{.752} & \centering{.710}  & \centering{.638} & \centering{.868} & \centering{.601} \tabularnewline
\emph{CRF only lexSyn} $\dagger$ & \centering{.762} & \centering{.780} & \centering{.744}  & \centering{.714} & \centering{.873} & \centering{.620} \tabularnewline
\emph{CRF only probability} & \centering{.605} & \centering{.698} & \centering{.534}  & \centering{.520} & \centering{.806} & \centering{.217} \tabularnewline
\hline
\emph{CRF w/o genre-dependent} $\dagger$ & \centering{.847} & \centering{.851} & \centering{.844} & \centering{\textbf{.778}} & \centering{.925} & \centering{.835} \tabularnewline
\emph{CRF all features} $\dagger$ & \centering{\textbf{.849}} & \centering{\textbf{.853}} & \centering{\textbf{.846}} & \centering{.777} & \centering{\textbf{.927}} & \centering{\textbf{.842}} \tabularnewline
	\end{tabularx}	
	}
\end{table} 
Syntactic features are useful for identifying the beginning of argument components. 
The probability feature yields the lowest score. 
Nevertheless, we observe a significant decrease of $.028$ F1 score of ``Arg-B'' when evaluating the system without the probability feature.
We obtain the best results by using all features.
Since persuasive essays exhibit a particular paragraph structure which may not be present in other text genres (e.g. user-generated web discourse), we also evaluate the model without genre-dependent features (cf.~Table~\ref{tab:featuresIdentification}). 
This yields a macro F1 score of $.847$ which is only $.002$ less compared to the model with all features.

Table \ref{tab:modelSelectionComps} shows the model selection results of the classification model. 
Structural features are the only features which significantly outperform the heuristic baseline when used individually. 
They are the most effective features for identifying major claims.
The second-best features for identifying claims are discourse features.
With this knowledge, we can confirm the assumption that general discourse relations are useful for component classification (cf. Section \ref{compClass}). 
\begin{table}[!ht]
\caption{Argument component classification ($\dagger$ = significant improvement over baseline heuristic)}\label{tab:modelSelectionComps}
\footnotesize{
	\begin{tabularx}{\textwidth}{ p{0.23\textwidth} | p{0.07\textwidth} p{0.07\textwidth}  p{0.07\textwidth} | p{0.13\textwidth} p{0.11\textwidth} p{0.11\textwidth} }
	 &  \centering{\emph{F1}} & \centering{\emph{P}} & \centering{\emph{R}} & \centering{\emph{F1 MajorClaim}} & \centering{\emph{F1 Claim}} & \centering{\emph{F1 Premise}} 
\tabularnewline
\hline
\emph{Baseline majority} & \centering{.257} & \centering{.209} & \centering{.333} & \centering{0} & \centering{0} & \centering{.771} \tabularnewline

\emph{Baseline heuristic} & \centering{.724} & \centering{.724} & \centering{.723} & \centering{.740} & \centering{.560} & \centering{\textbf{.870}} \tabularnewline
\hline

\emph{SVM only lexical} & \centering{.591} & \centering{.603} & \centering{.580} & \centering{.591} & \centering{.405} & \centering{.772} \tabularnewline
\emph{SVM only structural} $\dagger$ & \centering{.746} & \centering{.726} & \centering{.767} & \centering{.803} & \centering{.551} & \centering{\textbf{.870}} \tabularnewline
\emph{SVM only contextual} & \centering{.601} & \centering{.603} & \centering{.600} & \centering{.656} & \centering{.248} & \centering{.836} \tabularnewline
\emph{SVM only indicators} & \centering{.508} & \centering{.596} & \centering{.443} & \centering{.415} & \centering{.098} & \centering{.799} \tabularnewline
\emph{SVM only syntactic} & \centering{.387} & \centering{.371} & \centering{.405} & \centering{.313} & \centering{0} & \centering{.783} \tabularnewline
\emph{SVM only probability} & \centering{.561} & \centering{.715} & \centering{.462} & \centering{.448} & \centering{.002} & \centering{.792} \tabularnewline
\emph{SVM only discourse} & \centering{.521} & \centering{.563} & \centering{.484} & \centering{.016} & \centering{.538} & \centering{.786} \tabularnewline
\emph{SVM only embeddings} & \centering{.588} & \centering{.620} & \centering{.560} & \centering{.560} & \centering{.355} & \centering{.815} \tabularnewline
\hline
\emph{SVM all w/o prob} \& emb $\dagger$ & \centering{.771} & \centering{.771} & \centering{\textbf{.772}} & \centering{.855} & \centering{\textbf{.596}} & \centering{.863} \tabularnewline
\emph{SVM w/o genre-dependent} & \centering{.742} & \centering{.745} & \centering{.739} & \centering{.819} & \centering{.560} & \centering{.847} \tabularnewline
\emph{SVM all features} $\dagger$ & \centering{\textbf{.773}} & \centering{\textbf{.774}} & \centering{.771} & \centering{\textbf{.865}} & \centering{.592} & \centering{.861} \tabularnewline
	\end{tabularx}	
	}
\end{table} 
However, embedding features do not perform as well as lexical features. They yield lower F1 scores for major claims and claims.
Contextual features are effective for identifying major claims since they implicitly capture if an argument component is present in the introduction or conclusion (cf. Section \ref{compClass}).
Indicator features are most effective for identifying major claims but contribute only slightly to the identification of claims. 
Syntactic features are predictive of major claims and premises but are not effective for recognizing claims. 
The probability features are not informative for identifying claims, probably because forward indicators may also signal inner premises in serial structures. 
Omitting probability and embedding features yields the best accuracy. 
However, we select the best performing system by means of the macro F1 score which is more appropriate for imbalanced data sets.
Accordingly, we select the model which uses all features (Table \ref{tab:modelSelectionComps}).

The model selection results for relation identification are shown in Table~\ref{tab:modelSelectionRel}. 
We report the results of feature ablation tests since none of the feature groups yields remarkable results when used individually. 
\begin{table}[!ht]
\caption{Argumentative relation identification ($\dagger$ = significant improvement over baseline heuristic; $\ddagger$ = significant difference compared to SVM all features)}\label{tab:modelSelectionRel}
\footnotesize{
	\begin{tabularx}{\textwidth}{ p{0.26\textwidth} | p{0.09\textwidth} p{0.09\textwidth}  p{0.09\textwidth} | p{0.13\textwidth} p{0.13\textwidth} }
	 & \centering{\emph{F1}} & \centering{\emph{P}} & \centering{\emph{R}} & \centering{\emph{F1 Not-Linked}} & \centering{\emph{F1 Linked}} 
\tabularnewline
\hline
\emph{Baseline majority} & \centering{.455} & \centering{.418} & \centering{.500} & \centering{.910} & \centering{0}  \tabularnewline
\emph{Baseline heuristic} & \centering{.660} & \centering{.657} & \centering{.664} & \centering{.885} & \centering{.436}  \tabularnewline
\hline
\emph{SVM all w/o lexical} $\dagger$ & \centering{\textbf{.736}} & \centering{.762} & \centering{.711} & \centering{\textbf{.917}} & \centering{\textbf{.547}}  \tabularnewline
\emph{SVM all w/o syntactic} $\dagger$ & \centering{.729} & \centering{\textbf{.764}} & \centering{.697} & \centering{\textbf{.917}} & \centering{.526}  \tabularnewline
\emph{SVM all w/o structural} $\dagger$ & \centering{.715} & \centering{.740} & \centering{.692} & \centering{.911} & \centering{.511}  \tabularnewline
\emph{SVM all w/o indicators} $\dagger$ & \centering{.719} & \centering{.743} & \centering{.697} & \centering{.912} & \centering{.520}  \tabularnewline 
\emph{SVM all w/o discourse} $\dagger$ & \centering{.732} & \centering{.755} & \centering{.709} & \centering{.915} & \centering{.540}  \tabularnewline
\emph{SVM all w/o pmi} $\dagger$ & \centering{.720} & \centering{.745} & \centering{.697} & \centering{.912} & \centering{.521}   \tabularnewline
\emph{SVM all w/o shNo} $\dagger$ & \centering{.733} & \centering{.756} & \centering{\textbf{.712}} & \centering{.915} & \centering{.545}  \tabularnewline
\hline
\emph{SVM w/o genre-dependent} $\dagger$ & \centering{.722} & \centering{.750} & \centering{.700} & \centering{.913} & \centering{.520}  \tabularnewline
\emph{SVM all features} $\dagger$ & \centering{.733} & \centering{.756} & \centering{.711} & \centering{.915} & \centering{.544} \tabularnewline
	\end{tabularx}	
	}
\end{table} 
We also found that removing any of the feature groups does not yield a significant difference compared to the model with all features. 
Structural features are the most effective features for identifying relations. 
The second- and third-most effective feature groups are indicator and PMI features.
Both syntactic and discourse features yield a slight improvement when combining them with other features. 
Removing the shared noun features does not yield a difference in accuracy or macro F1 score although we observe a decrease of $.002$ macro F1 score when removing them from our best performing model. 
We achieve the best results by removing lexical features from the feature set. 

Table~\ref{tab:modelSelectionILP} shows the model selection results of the ILP joint model. 
\emph{Base+heuristic} shows the result of applying the baseline to all paragraphs in which the base classifiers identify neither claims nor argumentative relations. 
The heuristic baseline is triggered in $31$ paragraphs which results in $3.3\%$ more trees identified compared to the base classifiers. 
However, the difference between Base+heuristic and the base classifiers is not statistically significant. 
For this reason, we can attribute any further improvements to the joint modeling approach. 
\begin{table}[!ht]
\footnotesize
\setlength{\tabcolsep}{.1em}
\def\arraystretch{1.2}
\caption{Joint modeling approach ($\dagger$ = significant improvement over base heuristic; $\ddagger$ = significant improvement over base classifier; Cl$\rightarrow$Pr = number of claims converted to premises; Pr$\rightarrow$Cl = number of premises converted to claims; Trees = Percentage of correctly identified trees)}\label{tab:modelSelectionILP}
	\begin{tabularx}{\textwidth}{ p{0.14\textwidth} | p{0.03\textwidth} p{0.03\textwidth} p{0.03\textwidth}| p{0.02\textwidth} p{0.06\textwidth}  p{0.065\textwidth}  p{0.065\textwidth}  p{0.06\textwidth} | p{0.02\textwidth} p{0.06\textwidth} p{0.08\textwidth}  p{0.065\textwidth} |p{0.07\textwidth}  p{0.07\textwidth}  p{0.065\textwidth} }
	 & \multicolumn{3}{c|}{\emph{Parameter}} &	\multicolumn{5}{c|}{\emph{Components}} & \multicolumn{4}{c|}{\emph{Relations}} & \multicolumn{3}{c}{\emph{Statistics}}  \tabularnewline
	 & \centering{$\phi_r$} & \centering{$\phi_{cr}$} & \centering{$\phi_c$}&& \centering{F1} & \centering{F1 MC} & \centering{F1 Cl} & \centering{F1 Pr}  & & \centering{F1}  & \centering{F1 NoLi} & \centering{F1 Li} & \centering{Cl$\rightarrow$Pr} & \centering{Pr$\rightarrow$Cl} & \centering{Trees} \tabularnewline
	 \hline
	 \emph{Base heuristic} &\centering{-} &\centering{-}&\centering{-}& & \centering{.724} & \centering{.740} & \centering{.560} & \centering{.870} & & \centering{.660} & \centering{.885} & \centering{.436} & \centering{-} & \centering{-} & \centering{100\%} \tabularnewline
	  \emph{Base classifier} &\centering{-} &\centering{-}&\centering{-}& \centering{$\dagger$} & \centering{.773} & \centering{\textbf{.865}} & \centering{.592} & \centering{.861} & \centering{$\dagger$} & \centering{.736} & \centering{.917} & \centering{.547} & \centering{-} & \centering{-} & \centering{20.9\%} \tabularnewline
	\hline
	 \emph{Base+heuristic} &\centering{-} &\centering{-}&\centering{-}& \centering{$\dagger$} & \centering{.776} & \centering{\textbf{.865}} & \centering{.601} & \centering{.861} & \centering{$\dagger$} & \centering{.739} & \centering{.917} & \centering{.555} & \centering{0} & \centering{31} & \centering{24.2\%} \tabularnewline 
	\hline
	 \emph{ILP-na\"{i}ve} &\centering{$1$} &\centering{$0$} & \centering{$0$} & \centering{$\dagger$} & \centering{.765} & \centering{\textbf{.865}} & \centering{.591} & \centering{.761} & \centering{$\dagger$} & \centering{.732} & \centering{.918} & \centering{.530} & \centering{206} & \centering{1{,}144} & \centering{100\%} \tabularnewline
	 \emph{ILP-relation} &\centering{$\frac{1}{2}$} &\centering{$\frac{1}{2}$}&\centering{0}& \centering{$\dagger\ddagger$} & \centering{.809} & \centering{\textbf{.865}} & \centering{.677} & \centering{.875} & \centering{$\dagger\ddagger$} & \centering{\textbf{.759}} & \centering{\textbf{.919}} & \centering{\textbf{.598}} & \centering{299} & \centering{571} & \centering{100\%} \tabularnewline
	  \emph{ILP-claim} &\centering{$0$} &\centering{$0$}&\centering{$1$}& \centering{$\dagger$} & \centering{.740} & \centering{\textbf{.865}} & \centering{.549} & \centering{.777} &  & \centering{.666} & \centering{.894} & \centering{.434} & \centering{229} & \centering{818} & \centering{100\%} \tabularnewline
	\hline
	 \emph{ILP-equal} &\centering{$\frac{1}{3}$} &\centering{$\frac{1}{3}$}&\centering{$\frac{1}{3}$}& \centering{$\dagger\ddagger$} & \centering{.822} & \centering{\textbf{.865}} & \centering{.699} & \centering{.903} & \centering{$\dagger$} & \centering{.751} & \centering{.913} & \centering{.590} & \centering{294} & \centering{280} & \centering{100\%} \tabularnewline
	 \emph{ILP-same} &\centering{$\frac{1}{4}$} &\centering{$\frac{1}{4}$}&\centering{$\frac{1}{2}$}& \centering{$\dagger\ddagger$} & \centering{.817} & \centering{\textbf{.865}} & \centering{.687} & \centering{.898} & \centering{$\dagger$} & \centering{.738} & \centering{.908} & \centering{.569} & \centering{264} & \centering{250} & \centering{100\%} \tabularnewline
	 \emph{ILP-balanced} &\centering{$\frac{1}{2}$} &\centering{$\frac{1}{4}$}&\centering{$\frac{1}{4}$}& \centering{$\dagger\ddagger$} & \centering{\textbf{.823}} & \centering{\textbf{.865}} & \centering{\textbf{.701}} & \centering{\textbf{.904}} & \centering{$\dagger$} & \centering{.752} & \centering{.913} & \centering{.591} & \centering{297} & \centering{283} & \centering{100\%} \tabularnewline
	\end{tabularx}	
\end{table} 
Moreover, Table \ref{tab:modelSelectionILP} shows selected results of the hyperparameter tuning of the ILP joint model. 
Using only predicted relations in the ILP-na\"{i}ve model does not yield an improvement over the base classifiers. 
ILP-relation uses only information from the relation identification base classifier. 
It significantly outperforms both base classifiers but converts a large number of premises to claims. 
The ILP-claim model uses only the outcomes of the argument component base classifier and improves neither component classification nor relation identification. 
All three models identify a relatively high proportion of claims compared to the number of claims in our training data. 
The reason for this is that  many weights in $W$ are $0$. 
Combining the results of both base classifiers yields a considerably more balanced proportion of component type conversions. 
All three models (ILP-equal, ILP-same, and ILP-balanced) significantly outperform the base classifier for component classification. 
We identify the best performing system by means of the average macro F1 score for both tasks. 
Accordingly, we select ILP-balanced as our best performing ILP joint model.

Table \ref{tab:modelSelectionStance} shows the model selection results for the stance recognition model. 
Using sentiment, structural and embedding features individually does not yield an improvement over the majority baseline. 
However, lexical, syntactic and discourse features yield a significant improvement over the heuristic baseline when used individually. 
Although lexical features perform best individually, there is no significant difference when removing them from the feature set. 
Since omitting any of the feature groups yields a lower macro F1 score, we select the model with all features as the best performing model.

\begin{table}[!ht]
\caption{Stance recognition ($\dagger$ = significant improvement over baseline heuristic; $\ddagger$ = significant difference compared to SVM all features)}\label{tab:modelSelectionStance}
\footnotesize{
	\begin{tabularx}{\textwidth}{ p{0.25\textwidth} | p{0.09\textwidth} p{0.09\textwidth}  p{0.09\textwidth} | p{0.13\textwidth} p{0.13\textwidth}  }
	 & \centering{\emph{F1}} & \centering{\emph{P}} & \centering{\emph{R}} & \centering{\emph{F1 Support}} & \centering{\emph{F1 Attack}} \tabularnewline
\hline
\emph{Baseline majority} & \centering{.475} & \centering{.452} & \centering{.500} & \centering{.950} & \centering{0} \tabularnewline
\emph{Baseline heuristic} & \centering{.521} & \centering{.511} & \centering{.530} & \centering{.767} & \centering{.173} \tabularnewline
\hline
\emph{SVM only lexical} $\dagger$  & \centering{.663} & \centering{.677} & \centering{.650} & \centering{.941} & \centering{.383} \tabularnewline
\emph{SVM only syntactic} $\dagger$ & \centering{.649} & \centering{.725} & \centering{.587} & \centering{.950} & \centering{.283} \tabularnewline
\emph{SVM only discourse} $\dagger$  & \centering{.630} & \centering{\textbf{.746}} & \centering{.546} & \centering{\textbf{.951}} & \centering{.169} \tabularnewline
\hline
\emph{SVM all w/o lexical} $\dagger$ & \centering{.696} & \centering{.719} & \centering{.657} & \centering{.948} & \centering{.439} \tabularnewline
\emph{SVM all w/o syntactic} $\dagger\ddagger$ & \centering{.687} & \centering{.691} & \centering{.684} & \centering{.941} & \centering{.433} \tabularnewline
\emph{SVM all w/o sentiment} $\dagger$ & \centering{.699} & \centering{.710} & \centering{.688} & \centering{.945} & \centering{.451} \tabularnewline
\emph{SVM all w/o structural} $\dagger$  & \centering{.698} & \centering{.710} & \centering{.686} & \centering{.946} & \centering{.449} \tabularnewline
\emph{SVM all w/o discourse} $\dagger\ddagger$ & \centering{.675} & \centering{.685} & \centering{.666} & \centering{.941} & \centering{.408} \tabularnewline
\emph{SVM all w/o embeddings} $\dagger$  & \centering{.692} & \centering{.703} & \centering{.682} & \centering{.944} & \centering{.439} \tabularnewline
\hline
\emph{SVM all features} $\dagger$  & \centering{\textbf{.702}} & \centering{.714} & \centering{\textbf{.690}} & \centering{.946} & \centering{\textbf{.456}} \tabularnewline
	\end{tabularx}	
	}
\end{table}

\appendixsection{Indicators}
\label{sec:appendixIndicators}

\setcounter{table}{0}
\renewcommand{\thetable}{C\arabic{table}}

Table \ref{tab:listOfIndicators} shows all of the lexical indicators we extracted from $30$ persuasive essays. The lists include $22$ forward indicators, $33$ backward indicators, $48$ thesis indicators and $10$ rebuttal indicators.

\begin{table}[!ht]
\caption{List of lexical indicators}\label{tab:listOfIndicators}
\footnotesize{
	\begin{tabularx}{\textwidth}{ | p{0.12\textwidth}  | p{0.81\textwidth} | }
		\hline
		\emph{\textbf{Category}} & \emph{\textbf{Indicators}} \tabularnewline
		\hline
		\parbox[t]{2mm}{\multirow{1}{*}{\emph{Forward (24)}}} & ``As a result'', ``As the consequence'', ``Because'', ``Clearly'', ``Consequently'', ``Considering this subject'', ``Furthermore'', ``Hence'', ``leading to the consequence'', ``so'', ``So'', ``taking account on this fact'', ``That is the reason why'', ``The reason is that'', ``Therefore'', ``therefore'', ``This means that'', ``This shows that'', ``This will result'', ``Thus'', ``thus'', ``Thus, it is clearly seen that'', ``Thus, it is seen'', ``Thus, the example shows''\tabularnewline
		\hline
		\parbox[t]{2mm}{\multirow{1}{*}{\emph{Backward (33)}}} & ``Additionally'', ``As a matter of fact'', ``because'', ``Besides'', ``due to'', ``Finally'', ``First of all'', ``Firstly'', ``for example'', ``For example'', ``For instance'', ``for instance'', ``Furthermore'', ``has proved it'', ``In addition'', ``In addition to this'', ``In the first place'', ``is due to the fact that'', ``It should also be noted'', ``Moreover'', ``On one hand'', ``On the one hand'', ``On the other hand'', ``One of the main reasons'', ``Secondly'', ``Similarly'', ``since'', ``Since'', ``So'', ``The reason'', ``To begin with'', ``To offer an instance'', ``What is more''\tabularnewline
		\hline
		\parbox[t]{2mm}{\multirow{1}{*}{\emph{Thesis (48)}}} & ``All in all'', ``All things considered'', ``As far as I am concerned'', ``Based on some reasons'', ``by analyzing both the views'', ``considering both the previous fact'', ``Finally'', ``For the reasons mentioned above'', ``From explanation above'', ``From this point of view'', ``I agree that'', ``I agree with'', ``I agree with the statement that'', ``I believe'', ``I believe that'', ``I do not agree with this statement'', ``I firmly believe that'', ``I highly advocate that'', ``I highly recommend'', ``I strongly believe that'', ``I think that'', ``I think the view is'', ``I totally agree'', ``I totally agree to this opinion'', ``I would have to argue that'', ``I would reaffirm my position that'', ``In conclusion'', ``in conclusion'', ``in my opinion'', ``In my opinion'', ``In my personal point of view'', ``in my point of view'', ``In my point of view'', ``In summary'', ``In the light of the facts outlined above'', ``it can be said that'', ``it is clear that'', ``it seems to me that'', ``my deep conviction'', ``My sentiments'', ``Overall'', ``Personally'', ``the above explanations and example shows that'', ``This, however'', ``To conclude'', ``To my way of thinking'', ``To sum up'', ``Ultimately''\tabularnewline
		\hline
		\parbox[t]{2mm}{\multirow{1}{*}{\emph{Rebuttal (10)}}} & ``Admittedly'', ``although'', ``Although'', ``besides these advantages'', ``but'', ``But'', ``Even though'', ``even though'', ``However'', ``Otherwise''\tabularnewline
		\hline
	\end{tabularx}
	}
\end{table}

\newpage
\starttwocolumn

\begin{acknowledgments}

This work has been supported by the Volkswagen Foundation as part of the Lichtenberg-Professorship Program under grant No. I/82806 and by the German Federal Ministry of Education and Research (BMBF) as a part of the Software Campus project AWS under grant No. 01|S12054. We would like to thank the anonymous reviewers for their valuable feedback, 
  Can Diehl, Ilya Kuznetsov, Todd Shore and Anshul Tak for their valuable contributions, and Andreas Peldszus for providing details about his corpus. 

\end{acknowledgments}

\bibliographystyle{fullname}
\bibliography{CoLi2015}

\end{document}